\documentclass[conference]{IEEEtran}
\IEEEoverridecommandlockouts

\usepackage{cite}
\usepackage{amsmath,amssymb,amsfonts}
\usepackage{algorithmic}
\usepackage{graphicx}
\usepackage{textcomp}
\usepackage{float}  
\usepackage{subcaption}
\usepackage{booktabs}
\usepackage{multirow}
\usepackage{url}

\usepackage{xcolor}
\usepackage{placeins} 
\def\BibTeX{{\rm B\kern-.05em{\sc i\kern-.025em b}\kern-.08em
    T\kern-.1667em\lower.7ex\hbox{E}\kern-.125emX}}

\begin{document}

\title{Noise-Aware Misclassification Attack Detection in Collaborative DNN Inference\thanks{This material is based upon work supported by the National Science Foundation (NSF) under Award Number CNS-2401928.}
}

\author{Shima Yousefi, Saptarshi Debroy\\ City University of New York\\ Emails: \textit{syousefi@gradcenter.cuny.edu, saptarshi.debroy@hunter.cuny.edu}
\thanks{This work has been accepted for publication in IEEE/ACM CCGrid 2026. The final version will be available via IEEE Xplore.}
}

\maketitle

\begin{abstract}


Collaborative inference of object classification Deep neural Networks (DNNs) where resource-constrained end-devices offload partially processed data to remote edge servers to complete end-to-end processing, is becoming a key enabler of edge-AI. However, such edge-offloading is vulnerable to malicious data injections leading to stealthy misclassifications that are tricky to detect, especially in the presence of environmental noise.
In this paper, we propose a semi-gray-box and noise-aware anomaly detection framework fueled by a variational autoencoder (VAE)  to capture deviations caused by
adversarial manipulation. The proposed framework incorporates a robust noise-aware feature that captures the characteristic behavior of environmental noise to improve detection accuracy while reducing  false alarm rates. Our evaluation with popular object classification DNNs demonstrate the robustness of the proposed detection (up to 90\% AUROC across DNN configurations) under realistic noisy conditions while revealing limitations
caused by feature similarity and elevated noise levels.

\end{abstract}


\section{Introduction}
In recent years, deep neural networks (DNNs) are increasingly integrated into a wide range of Internet of Things (IoT) applications that enhance our daily lives, including smart cities, healthcare monitoring, and emergency response~\cite{wu2024deep,effect-dnn,vec}. End-devices often generate massive amounts of data and require rapid DNN inference with low latency to support real-time control decisions. However, performing DNN inference directly on end-devices remains a significant challenge due to fundamental hardware limitations.
A promising solution to address these limitations is collaborative inference, which distributes the computational workload involved in DNN inference between end-devices and remote edge/cloud servers \cite{zhang2025c2mec,xiaojie-sec}. In this paradigm, the DNN inference task is partitioned between the end-device and the edge server based on the available computational and energy resources, and the partition DNN layer (a.k.a. cut-point layer), 
where the end-device executes the initial set of layers (till partition) and then transmits the resulting intermediate feature representations, along with the cut-point information, to the edge server. The server then executes the remaining layers of the model to complete the DNN inference task \cite{inferedge}.
Therefore, the success of such 
collaborative inference depends on the integrity of transmitted intermediate features from the end-devices to servers.

However, such data exchange often uses unsecured wireless links or intermediate devices/hops. This makes such collaborative inference vulnerable to two primary threats: (i) Data privacy  \cite{he2019model}, and (ii) Data integrity \cite{yousefi2025advar}.
While prior studies have extensively addressed privacy leakage in collaborative inference and proposed solutions to mitigate such risks \cite{he2020attacking}, the integrity aspect, specifically how adversaries can alter intermediate data to mislead final decisions, and how they can be detected, is important and so far has been underexplored. Among integrity-oriented threats, the misclassification attack is particularly critical, where adversaries aim to manipulate transmitted data so that the edge-side model produces incorrect predictions. Most existing works on adversarial misclassification assume that attackers perturb the raw input data, for example, by modifying a single pixel in an image to induce misclassification \cite{su2019one} before it is processed by the DNN.

What makes detecting and defending against such confidentiality and integrity threats non-trivial in edge-native collaborative inference is the well-known energy constraints of most resource-constrained end-devices. This makes implementing conventional security mechanisms, such as fully homomorphic encryption and/or secure multiparty computation (SMC) burdensome, since they introduce significant computational and power overhead \cite{mireshghallah2020principled}. Likewise, adopting differential privacy for intermediate processed data is often infeasible, as it changes the intermediate feature representations and degrades the accuracy of downstream inference tasks at the edge server \cite{he2020attacking}. In addition, 
the use cases that employ such collaborative inference, rely on sensor-enabled (e.g., camera) end-devices that operate under inherently noisy conditions due to communication errors, thermal fluctuations, hardware aging, and other environmental factors. 
The presence of such noise can distort or shift feature representations before they can reach the edge, similar to how industrial IoT systems encounter various forms of noise that corrupt time-series data \cite{liu2020noise}. Such natural noise may cause clean data to resemble adversarial behavior and blur the decision boundary used by state-of-the-art detection methods, thereby making the design of an effective detection strategy non-trivial.
In this paper, we focus on the aforementioned system vulnerabilities of collaborative DNN inference. Specifically, we explore the detection of stealthy misclassification attacks in noisy edge-native collaborative DNN inference environments. We explicitly consider the effect of environmental noise on the exchange of intermediate feature representations. We focus on intermediate features rather than raw inputs, as adversarial manipulation in collaborative inference directly targets these representations to induce incorrect downstream predictions.
To address this challenge, we propose a 
{\em semi–gray-box} noise-aware detection framework. The framework operates in a {\em semi–gray-box} setting in which structured prior knowledge of the underlying stochastic noise process affecting intermediate feature representations is available to the detector, while the internal parameters of the DNN and the adversarial generation mechanism remain unknown. We consider a state-of-the-art black-box adversarial attack that leverages a variational autoencoder (VAE)~\cite{yousefi2025advar}, which manipulates transmitted feature representations to induce incorrect edge-server predictions.
To detect such attack, we propose a VAE-based representation to capture structural deviations caused by adversarial manipulation. In noisy environments, benign stochastic variations may overlap with adversarial perturbations, reducing the separability between benign and manipulated representations. To mitigate this, we incorporate a robust noise-aware feature that captures the characteristic behavior of noise and reduces false alarms.
By operating solely on observable intermediate features and compact noise-aware descriptors, the proposed detection leverages prior structure in noise behavior without access to internal model parameters. Consequently, the detection can be deployed online with minimal overhead during operational period, requiring no additional model retraining. 


We evaluate the effectiveness of the proposed detection method across multiple state-of-the-art convolutional neural network (CNNs) architectures, including VGG19, AlexNet, and MobileNet, under varying noise intensities and an optimal misclassification attack scenario in which the adversary maximizes misclassification confidence. Detection performance is measured based on detection accuracy in terms of the ability to separate adversarial samples from benign ones in the feature space. 
The results show that the proposed detection framework achieves strong separability under a wide range of noise conditions with detection performance exceeding 80-90\% AUROC across multiple cut-point layer configurations. The findings indicate that detection performance is layer-dependent and degrades as adversarial representations become more similar to benign features at deeper cut-points. Additionally, increasing noise intensity leads to greater overlap in the feature space, making detection more challenging and causing performance to approach near-random levels. Overall, our results demonstrate the robustness of the proposed detection under realistic noisy conditions while revealing limitations caused by feature similarity and elevated noise levels.


The rest of the paper is organized as follows. Section \ref{sec:Related_Work} reviews related work and background. Section \ref{sec:motivation} presents the system model, threat model, and motivating examples. Section \ref{sec:detection} discusses the proposed detection methodology.
Section \ref{sec:evaluation} describes the evaluation setup and results. Section \ref{sec:conclusion} concludes the paper.

\section{Related Work}
\label{sec:Related_Work}

As end-devices possess limited computational power and energy resources, it is impractical for most of them to train DNN models locally. Instead, they typically rely on pre-trained models, which are deployed collaboratively between the devices and edge server(s) to enable efficient inference. However, such pre-trained models are vulnerable to inference-time or evasion attacks, where adversaries, without access to the training process, exploit the model to induce misclassification or misdetection~\cite{biggio2013evasion}. In addition, attackers can leverage model outputs or intermediate representations to reconstruct sensitive input data or extract model parameters~\cite{he2019model}.
Data manipulation and leakage are also some of the most critical security risks across the different layers of the end-device and edge server collaborative environments~\cite{saha2025detection, man2020ghostimage}. Recent studies have demonstrated that data falsification attacks can significantly degrade system-level performance objectives, including energy efficiency and end-to-end latency in collaborative IoT–edge environments \cite{yousefi2024intent}.
Integrity violation during the DNN model inference phase aims to manipulate model predictions. DNN models are known to be highly vulnerable to adversarial examples, where an adversary introduces small, carefully crafted perturbations, often imperceptible to the human eye, into the input data \cite{gan2025exploring}. These modifications can cause the model to produce incorrect predictions or misclassify inputs into attacker-specified target classes that differ from the ground truth~\cite{guetta2021dodging}.
Anomaly detection mechanisms across different application domains, including intrusion detection systems that leverage machine learning to reduce false alerts \cite{mahboob2021coronavirus}, are categorized based on their purpose and the stage at which they are applied.
The training phase detections aim to enhance a model’s robustness by modifying the learning algorithm or incorporating adversarial samples into the training process \cite{goodfellow2014explaining}. For example, recent multimodal generative frameworks leverage diffusion-based augmentation to mitigate class imbalance in intrusion detection datasets\cite{loodaricheh2026mage}. However, these approaches are less practical for latency-sensitive applications. In contrast, inference stage detections operate without retraining the target model and are more suitable for IoT–edge environments. For instance, \cite{feinman2017detecting} proposes an inference-stage adversarial detection method based on Bayesian uncertainty estimates and statistical deviations from normal data. However, the required additional inference-time processing introduces computational overhead, making it less suitable for resource-constrained IoT–edge systems. Authors in \cite{qing2024detection} propose ADDNP, a detection-based method that leverages reconstruction discrepancy (RD) in the learned latent feature space. The method exploits the observation that adversarial inputs exhibit higher intrinsic dimensionality than normal samples, leading to significantly larger reconstruction errors that enable effective adversarial detection. Authors in \cite{mu2025robust} propose a detection method based on high-level feature differences. The method uses an encoder to measure the similarity between a test input and a randomly selected reference sample with the same predicted label. If the feature similarity is low, the input is flagged as an adversarial example.

Existing adversarial detection techniques typically assume clean transmission of model inputs or intermediate representations. However, collaborative IoT–edge environments, especially those supporting mission-critical applications, operate under inherently noisy conditions. Such noise may originate from wireless communication channels, environmental thermal fluctuations, hardware degradation, or other operational perturbations within the IoT–edge system. Noisy samples are unavoidable in real-world applications. They can distort the boundaries between normal and abnormal behavior, making it harder to distinguish benign noise-affected data from truly adversarial samples \cite{bigdeli2017fast}. As a result, it becomes essential to account for the behavior of such noise when designing reliable adversarial detection methods.


\section{System and Threat Model}
\label{sec:motivation}
 
\subsection{System Model \& Motivating Example}
The system considered in this work is a collaborative end-device and edge inference framework in which the end-device executes the early layers of a DNN and transmits the resulting intermediate feature representations to the edge server. The edge server then processes the remaining layers. As discussed earlier, this architecture introduces a network communication step where noise may distort the transmitted features. Importantly, moderate levels of noise do not necessarily degrade the downstream DNN’s classification accuracy, as modern models are often robust to small perturbations. However, such noise can significantly affect detection mechanisms that rely on the geometry of the feature space (e.g., reconstruction error, latent shift). 
Studies have shown that DNNs are generally robust to small perturbations in the input image domain \cite{ momeny2021noise}. However, it is not clear whether this robustness extends to intermediate representations used in collaborative inference, where noise is injected after the early layers (i.e., till the cut-point). To examine this, 
we retrain a widely used CNN, viz., VGG19 on the CIFAR-100 dataset and use its 10,000-sample test set to evaluate robustness under noise. For each test input, we extract the intermediate representation at layer 20 and inject noise with the parameters shown in Tab.\ref{tab:noise_robustness}.
The selected parameter values represent increasing levels of impulsive interference. The injected noise follows the symmetric alpha-stable (S$\alpha$S) distribution described in Section \ref{sec:noise-model}. In the experiments, we consider symmetric ($\eta = 0$) zero-mean ($\delta = 0$) noise. The parameter $\alpha\in(0,2]$ determines the impulsiveness of the noise, while the parameter $\kappa$ controls the scale of the noise. In addition to these distribution-level parameters, we introduce control parameters in the experimental setup, including a corruption probability $p_b$ and a fraction of noisy samples $f_n$, to simulate communication disturbances.
The noisy features are then forwarded through the remaining layers of VGG19 at the edge server. The table reports the end-to-end classification accuracy and the change in Right–Wrong Confidence Gap (RWCG) under increasing noise intensities. As shown, light, moderate, and even severe noise levels yield negligible accuracy degradation, confirming that the downstream classifier remains stable even when intermediate features are perturbed. Below, we describe the noise model used in this work and present a motivating example that illustrates why incorporating noise-awareness into adversarial detection is essential for these systems.

\begin{table}[t]
\centering
\scriptsize
\setlength{\tabcolsep}{4.5pt}   
\renewcommand{\arraystretch}{1.1}
\caption{
Robustness of the downstream VGG19 classifier under different noise levels. 
}
\label{tab:noise_robustness}
\begin{tabular}{lccccc c}
\toprule
Level & $\alpha$ & $\kappa$ & $p_b$ & $f_n$ & $\Delta$Acc(\%) & $\Delta$RWCG \\
\midrule
No noise & --  & --    & --    & 0.00 & 0.00  & 0.000 \\
Light    & 1.8 & 0.01  & 0.005 & 0.15 & 0.00  & 0.000 \\
Moderate & 1.6 & 0.02  & 0.010 & 0.30 & -0.16 & +0.011 \\
Severe   & 1.4 & 0.08  & 0.10  & 0.30 & -0.57 & -0.006 \\
Extreme  & 1.2 & 0.12  & 0.15  & 0.50 & -8.29 & -0.054 \\
\bottomrule
\end{tabular}
\vspace{-0.2in}
\end{table}

\subsubsection{Noise model}
\label{sec:noise-model}
Real-world end-device-edge deployments often operate in environments with significant noise. Prior studies in industrial settings showed that such noise is frequently electromagnetic in nature, generated from temperature variations, mechanical vibrations, and interference from surrounding machinery \cite{li2019measurement}. These disturbances degrade wireless communication quality and reduce the reliability of data transmission in low-power end-devices. Also, in systems equipped with imaging sensors, the raw data is inherently noisy due to the physical characteristics of digital sensors. In such sensors, part of the variability originates from photon statistics, making the noise signal-dependent. Accurately modeling this behavior is essential for effective image preprocessing and has been shown to play an important role in many imaging and computer vision applications \cite{foi2008practical}. In digital imaging, noise is often described through simple mathematical models. 
Many disturbances are additive, expressed as:
\[
f(x) = g(x) + q(x),
\]
where \(g(x)\) denotes the ideal noise-free signal and \(q(x)\) represents the noise component added to it. While other types of noise models exist, such as multiplicative, the noise introduced by the communication channel is commonly modeled as additive \cite{lapidoth2020encoder}.

In collaborative inference, intermediate feature representations produced by the end-device must be transmitted to the edge server, often over unsecured networks. Noise in the communication link or device hardware perturbs these features, shifting their position in latent space. Unlike anomalies, which are caused by abnormal or malicious behavior, noise distorts the structure of benign data, making reliable detection more difficult. This distortion increases false positives (i.e., noisy but normal samples incorrectly flagged as anomalies), which is the primary challenge for existing detection methods.

To capture this effect, we adopt an additive impulsive noise model based on the 
symmetric alpha-stable (S$\alpha$S) distribution, which is widely used to represent 
heavy-tailed interference in communication systems \cite{ma2012performance}. Since alpha-stable distributions do not have closed-form probability density functions for most values of $\alpha$, they are typically defined through their characteristic functions. In this work, the characteristic function of the 
noise variable $N$ is written as: 

\[
\Phi_{N}(u)
= \exp\!\left\{\, i\delta u 
- \kappa |u|^{\alpha}
\left[\, 1 + i\,\eta\,\text{sgn}(u)\,\Omega(u,\alpha)\, \right] 
\right\},
\]

where $\alpha\in(0,2]$ controls the impulsiveness of the noise, $\eta$ is the 
symmetry parameter, $\kappa$ is a scale factor, and $\delta$ is the location parameter.

The function $\Omega(u,\alpha)$ is defined as
\[
\Omega(u,\alpha) = 
\begin{cases}
\tan\!\left(\dfrac{\pi \alpha}{2}\right), & \alpha \neq 1, \\[10pt]
\dfrac{2}{\pi}\,\log|u|, & \alpha = 1,
\end{cases}
\]
and the sign function is given by
\[
\text{sgn}(u) = 
\begin{cases}
1, & u>0,\\
0, & u=0,\\
-1, & u<0.
\end{cases}
\]

Here, $N$ denotes the scalar random variable modeling the impulsive communication noise added to each dimension of the transmitted intermediate feature vector. In this work, we model the received intermediate feature vector as
\[
\tilde{\mathbf{z}} = \mathbf{z} + \mathbf{n},
\]
where $\mathbf{z}$ denotes the clean intermediate features produced on the end device or by the adversary, 
$\mathbf{n}$ is the S$\alpha$S impulsive noise term, and $\tilde{\mathbf{z}}$ is the 
corrupted feature observed at the edge server.

\subsubsection{Motivating example}
To illustrate the impact of communication noise on adversarial example detection, we consider a collaborative inference scenario based on an existing prior attack model. At the end-device side, intermediate feature representations extracted at the cut-point layer are either: (i) benign features produced by clean inputs or (ii) adversarial features crafted by an attack VAE to induce misclassification at the edge. We assume that the system has an adVAE-based detection trained solely on benign intermediate features, under the realistic assumption that, during an initial attack-free period, the system has access to clean data for establishing a normal reference distribution. The system uses reconstruction error and latent-space deviation to distinguish benign features from manipulated ones. Below, we discuss the accuracy of such detection to demonstrate detection challenges under noisy conditions.

\begin{figure}[t]
    \centering

    \includegraphics[width=0.8\columnwidth]{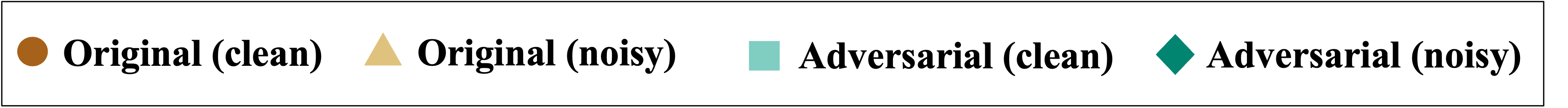}
    \vspace{0.5em}

    \begin{subfigure}[b]{0.49\columnwidth}
        \centering
        \includegraphics[width=\linewidth]
        {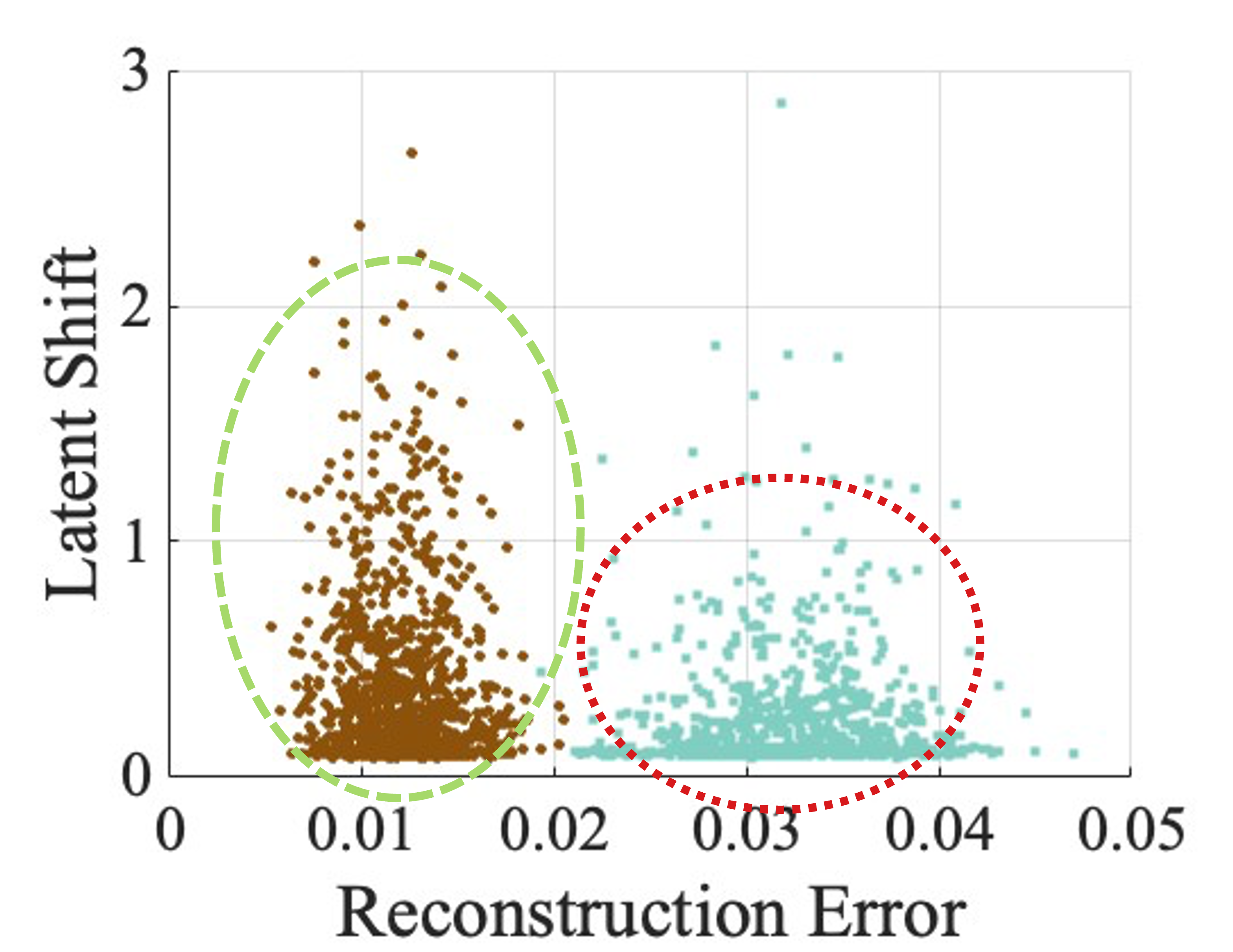} 
        \caption{\footnotesize Clean environment}
        \label{fig:clean_env}
    \end{subfigure}
    \hfill
    \begin{subfigure}[b]{0.49\columnwidth}
        \centering
        \includegraphics[width=\linewidth]{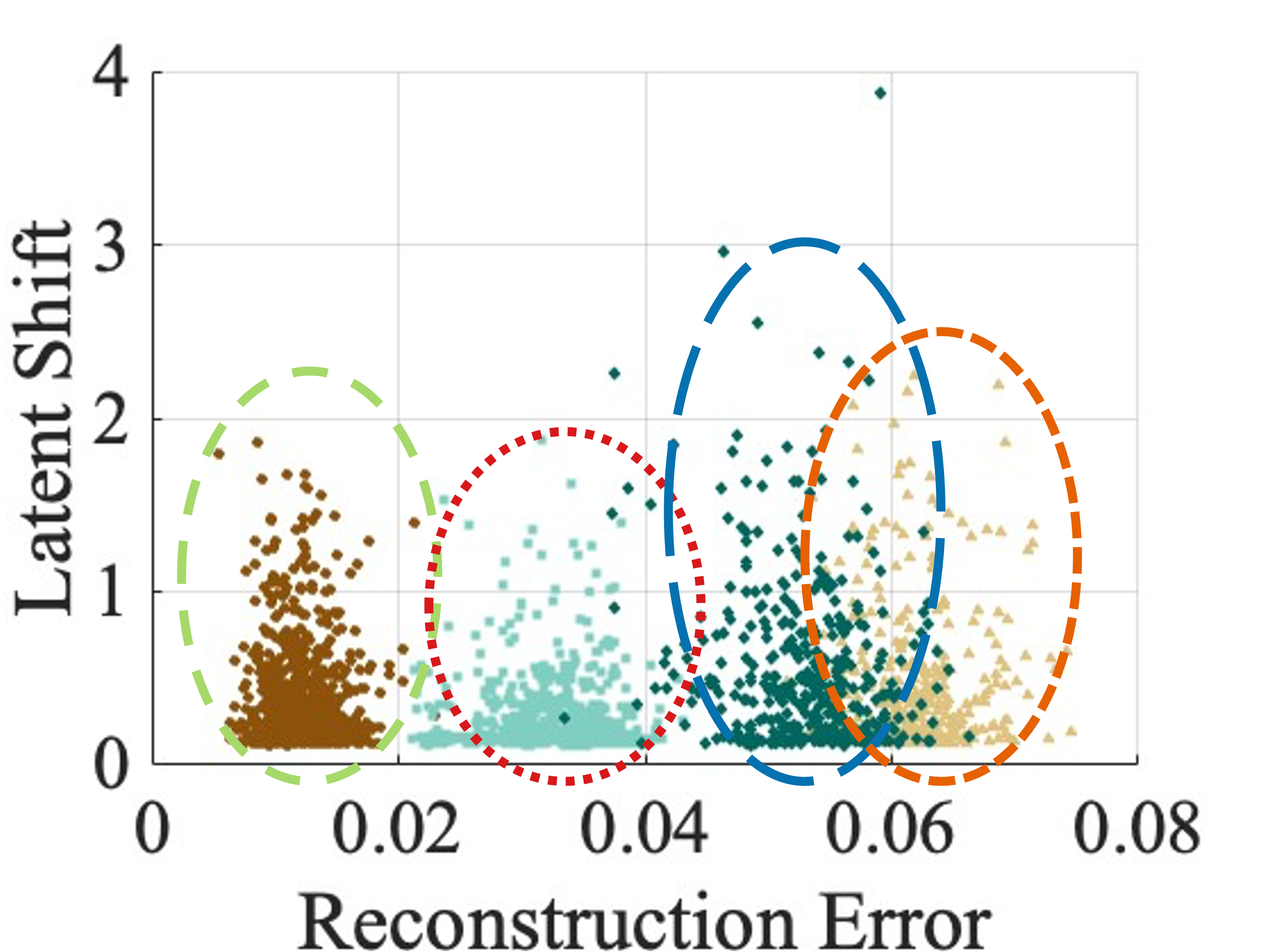} 
        \caption{\footnotesize Noisy environment}
        \label{fig:noisy_env}
    \end{subfigure}

    \caption{ \scriptsize Effect of noise on the distribution. Reconstruction error vs.\ latent shift for (a) clean intermediate features and (b) noisy S$\alpha$S-corrupted intermediate features.}
    \label{fig:clean_vs_noisy}
    \vspace{-0.2in}
\end{figure}

\noindent\underline{\textit{Detection Under Clean Environment:}} In this experiment, we evaluate a standard object classification model, VGG19, using feature representations extracted from the CIFAR-100 dataset. We construct an evaluation set containing a mixture of benign intermediate features and adversarial features generated by an attack VAE. These samples are passed through the adVAE-based detection, which computes two anomaly indicators: latent shift and reconstruction error. Fig. \ref{fig:clean_env} shows the scatter plot of latent shift versus reconstruction error. Under clean transmission conditions, benign and adversarial samples form well-separated clusters, allowing the detector to distinguish between the two clearly. This results in high detection accuracy and very low false-positive rates, demonstrating that the adVAE framework is effective when intermediate features are not corrupted by noise.

\noindent\underline{\textit{Detection Under Noisy Environment:}} Next, we evaluate detection performance when the same intermediate features are corrupted by the impulsive additive noise model described in Section~\ref{sec:noise-model}. As shown in Fig.~\ref{fig:noisy_env}, the presence of impulsive noise perturbs the latent-space representation of both benign and adversarial samples. This distortion shifts the anomaly scores and causes significant overlap between the two classes. Consequently, many noise-corrupted but benign samples are incorrectly flagged as adversarial, leading to elevated false positives and a substantial degradation in overall detection reliability.

To further quantify the impact of communication noise on adversarial detection, we compute reconstruction error and latent shift using an adVAE model. These two-dimensional anomaly scores are then evaluated using both a radius-based detector and a One-Class SVM. As shown in Tab.~\ref{tab:detector_comparison}, under noisy conditions, all methods experience a sharp degradation in their ability to draw a reliable boundary between benign and adversarial samples.
Although the detector performs well under clean conditions, simply denoising the received features before applying the detection mechanism is not straightforward. A denoising step introduces additional transformations in feature space, which may change the latent-space structure and affect the detector’s performance.
{\em This motivates the need to design noise-aware adversarial detection mechanisms, especially for safety-critical monitoring systems(e.g., industrial monitoring). Rather than treating every deviation from the clean latent manifold as an attack, a robust detection must incorporate additional noise-tolerant anomaly scores and explicitly model how noise affects the behavior of benign samples.}

\begin{table}[t]
\scriptsize
\centering
\caption{\scriptsize Anomaly detection performance under clean and noisy environments 
using different feature representations.}
\begin{tabular}{lcccccc}
\toprule
\textbf{Method} & \textbf{Env.} & \textbf{Acc.} & \textbf{Prec.} & 
\textbf{Rec.} & \textbf{F1} & \textbf{FPR} \\
\midrule

\multirow{2}{*}{adVAE\cite{wang2020advae} + Radius} 
& Clean & 0.969 & 0.996 & 0.945 & 0.970 & 0.004 \\
& Noisy & 0.835 & 0.778 & 0.958 & 0.859 & 0.301 \\
\midrule

\multirow{2}{*}{adVAE + OC-SVM} 
& Clean & 0.85 & 1.00 & 0.70 & 0.301 & 0.880 \\
& Noisy & 0.73 & 0.51 & 0.97 & 0.028 & 0.660 \\
\bottomrule
\end{tabular}
\label{tab:detector_comparison}
\vspace{-0.2in}
\end{table}

\subsection{Threat Model}
\label{sec:threat}
For the threat model, we assume a black-box adversary with no knowledge of the deployed DNN architecture, its parameters, the cut-point layer, or the original input data. The attacker’s capability is limited to intercept or access the intermediate feature representations generated at the end-device and transmitted to the edge server. 
Here, the adversary does not have access to the detection mechanism deployed at the edge server and cannot query the system to obtain anomaly scores or detection feedback. 

The attacker's objective is to manipulate the intermediate features so that the downstream DNN produces a high-confidence misclassification, thereby bypassing threshold-based anomaly detection or triggering mechanisms. Following the attack model described in \cite{yousefi2025advar}, we consider an attack strategy where the adversary learns to generate adversarial intermediate features that induce misclassification on the edge server after running the entire targeted DNN model. The workflow of the attack is as follows:
\begin{itemize}
    \item During passive observation, the adversary accumulates intermediate feature vectors transmitted from the end device to the edge server at the cut-point layer. This accumulated dataset of activations, denoted as \( \mathcal{D}_h = \{ h_1, h_2, \dots, h_N \} \), forms the basis for learning the structure of feature space. Prior analysis has demonstrated that these features enable an adversary to differentiate between models or cut-layer positions, even without knowledge of the model architecture or training data. This allows the adversary to organize the collected data before training the VAE. 
    \item Using the collected features $\mathcal{D}_h$, the adversary trains a VAE. The VAE encoder $g(\cdot)$ maps each intermediate feature $h_i$ into a latent code $z_i=g(h_i)$, creating a continuous latent space $\mathcal{Z}$ that captures the distribution of legitimate intermediate representations.
    \item To craft adversarial samples, the adversary then samples a latent vector $z_t \sim p_{\mathcal{Z}}$ from the learned latent distribution. This latent serves as a target representation that leads the downstream classifier toward a different output. During inference time, the benign intermediate feature $h_o$ generated by the end-device is encoded as $z_o = g(h_o)$. The adversary intercepts the benign latent representation $z_o$ and blends it with the target latent using linear interpolation:
    \[
        z_{\nu} = (1 - \nu) z_o + \nu z_t,
    \]
    where the interpolation coefficient $ \nu \in [0,1]$ is the attacker strength that determines how far the manipulated representation deviates from the original. Small  $\nu$ keeps the manipulated representation close to the benign latent code, making the attack more stealthy. While large $\nu$ pushes the latent toward the target direction and produces greater changes in the final DNN output.

    \item Finally, the modified latent vector  $z_{\nu}$  is passed through the VAE decoder to synthesize a new intermediate feature. This adversarial representation is transmitted to the edge server for final inference.
\end{itemize}

The effects of such attack on popular CNNs (e.g., VGG19, AlexNet, MobileNet) are shown in Tabs.~\ref {tab:alexnet_attack},~\ref{tab:VGG19_attack}, and \ref{tab:MobileNet_attack}.
For each model, the attacker applies different perturbation strengths $\nu$ at several cut-layer positions with different output sizes. 
Tab.~\ref {tab:alexnet_attack} presents the attack success rate (ASR) and the average confidence of AlexNet when subjected to adversarial intermediate features generated at different cut-point layers and for varying attack strengths $\nu$. The results reveal that attacks launched from deeper layers are significantly more effective than those originating from earlier layers, which primarily encode low-level features. As the cut-point layer moves deeper into the network, the intermediate representations become more semantically meaningful. This richer structure enables the VAE to synthesize more realistic adversarial features, which in turn drive AlexNet toward incorrect predictions with higher confidence. Increasing $\nu$ amplifies this effect, producing a sharp rise in ASR for deeper layers. However, very large $\nu$ values may introduce noticeable deviations that increase the likelihood of detection. This highlights that an attacker can tune ~$\nu$ to balance stealthiness against attack effectiveness.

The results of VGG19 are summarized in Tab.~\ref {tab:VGG19_attack}. As the cut-point layer moves deeper, the attack becomes substantially more effective, with ASR rising sharply as the attack strength~$\nu$ increases. This demonstrates that deeper layers encode richer semantic information, giving the VAE a more expressive latent structure to generate adversarial features. Additionally, VGG19 exhibits a clear confidence shift as the attack strength grows. This combination of high ASR and high confidence makes the attack particularly stealthy and difficult to detect using threshold-based mechanisms that rely solely on the final model output at the edge server.

Finally, Tab.~\ref{tab:MobileNet_attack} summarizes the attack results for the MobileNet model, which behaves differently from both VGG19 and AlexNet when exposed to adversarial intermediate representations. Unlike the other models, MobileNet is inherently uncertain toward VAE-generated samples and shows low classification confidence across all values of the attack strength $\nu$. However, despite this low confidence, the ASR remains high. This shows that MobileNet is easier to fool but less likely to remain stealthy for any confidence-based anomaly detector applied after the full inference pipeline.

While such a confidence-based triggering mechanism could catch many MobileNet adversarial samples, relying on post-inference detection is not ideal in collaborative inference environments.
Processing adversarial intermediate features at the server still consumes computational resources on a shared, resource-constrained edge node. More importantly, across the different models studied, including those like VGG19 that misclassify adversarial samples with high confidence.
These attacks pose a serious safety risk. Once manipulated features reach the server, the downstream DNN may produce confident but incorrect outputs, potentially triggering unsafe actions or control responses in mission-critical systems.
Therefore, adversarial samples must be identified before they are processed at the edge server. Early detection is essential not only for safety, but also for system efficiency. Even though the edge server in a collaborative inference setting is more powerful than end-devices, serving many devices simultaneously makes it resource-limited in practice. Wasting computation on malicious or meaningless tasks reduces the system’s overall capacity and increases latency for legitimate devices. These concerns motivate the development of a noise-aware adversarial detection mechanism capable of flagging harmful intermediate representations at the early stage.

\begin{table}[t]
\centering
\scriptsize
\setlength{\tabcolsep}{1.5pt}
\renewcommand{\arraystretch}{0.80}

\caption{\scriptsize AlexNet performance across layers and attack strengths (Acc removed).}
\label{tab:alexnet_attack}

\begin{tabular}{c|cc|cc|cc|cc}
\toprule
\multirow{2}{*}{$\nu$} &
\multicolumn{2}{c|}{\textbf{Layer 3 \%}} &
\multicolumn{2}{c|}{\textbf{Layer 6 \%}} &
\multicolumn{2}{c|}{\textbf{Layer 8 \%}} &
\multicolumn{2}{c}{\textbf{Layer 10 \%}} \\
\cline{2-9}
& \scriptsize ASR &\scriptsize  Conf 
& \scriptsize ASR & \scriptsize Conf 
& \scriptsize ASR & \scriptsize Conf 
& \scriptsize ASR & \scriptsize Conf \\
\midrule
0.0 & 83.33 &40.76 &43.57 & 52.82& 53.83& 46.79& 47.42& 64.26\\
0.2 & 91.02 &32.25 &57.68 & 47.64&65.37 &38.54 & 64.09& 54.43\\
0.4 & 96.15 & 27.23& 85.89& 31.66&83.33 &32.56 & 88.46& 54.23\\
0.6 &98.72 & 29.49& 96.15&34.97 & 94.87& 36.24& 97.43& 74.25\\
0.8 &97.43 &38.73 & 98.72& 58.86& 97.43& 49.22& 100.00& 91.44\\
1.0 & 96.15& 51.04& 98.72&82.26 & 100.00& 49.79&100.00 &96.65 \\
\bottomrule
\end{tabular}
\end{table}

\begin{table}[t]
\centering
\scriptsize
\setlength{\tabcolsep}{1.5pt}
\renewcommand{\arraystretch}{0.80}

\caption{\scriptsize VGG19 performance across layers and attack strengths (Acc removed).}
\label{tab:VGG19_attack}

\begin{tabular}{c|cc|cc|cc|cc}
\toprule
\multirow{2}{*}{$\nu$} &
\multicolumn{2}{c|}{\textbf{Layer 12 \%}} &
\multicolumn{2}{c|}{\textbf{Layer 16 \%}} &
\multicolumn{2}{c|}{\textbf{Layer 18 \%}} &
\multicolumn{2}{c}{\textbf{Layer 20 \%}} \\
\cline{2-9}
& \scriptsize ASR & \scriptsize Conf 
& \scriptsize ASR & \scriptsize Conf
& \scriptsize ASR & \scriptsize Conf
& \scriptsize ASR & \scriptsize Conf \\
\midrule
0.0 & 76.89 & 49.53 & 39.10 & 63.50 & 18.49 & 74.66 & 5.11 & 88.51 \\
0.2 & 82.97 & 42.18 & 41.73 & 63.50 & 35.52 & 70.62 & 48.91 & 82.87 \\
0.4 & 92.70 & 35.54 & 46.29 & 61.69 & 81.75 & 70.52 & 97.57 & 87.07 \\
0.6 & 97.57 & 39.42 & 48.21 & 66.41 & 100.00 & 82.28 & 91.48 & 95.74 \\
0.8 & 98.78 & 57.01 & 48.36 & 68.01 & 100.00 & 89.58 & 100.00 & 98.53 \\
1.0 & 98.78 & 59.63 & 48.36 & 68.58 & 100.00 & 93.92 & 100.00 & 98.79 \\
\bottomrule
\end{tabular}
\vspace{-0.2in}
\end{table}

\begin{table}[t]
\centering
\scriptsize
\setlength{\tabcolsep}{1.5pt}
\renewcommand{\arraystretch}{0.80}

\caption{\scriptsize MobileNet performance across layers and attack strengths (Acc removed).}
\label{tab:MobileNet_attack}

\begin{tabular}{c|cc|cc|cc|cc}
\toprule
\multirow{2}{*}{$\nu$} &
\multicolumn{2}{c|}{\textbf{Layer 20 \%}} &
\multicolumn{2}{c|}{\textbf{Layer 40 \%}} &
\multicolumn{2}{c|}{\textbf{Layer 50 \%}} &
\multicolumn{2}{c}{\textbf{Layer 63 \%}} \\
\cline{2-9}
& \scriptsize ASR & \scriptsize Conf
& \scriptsize ASR & \scriptsize Conf
& \scriptsize ASR & \scriptsize Conf
& \scriptsize ASR & \scriptsize Conf \\
\midrule
0.0 & 97.12 & 51.38 & 94.24 & 81.13 & 77.67 & 55.22 & 88.70 & 48.81 \\
0.2 & 98.56 & 46.40 & 97.12 & 81.91 & 94.24 & 49.66 & 88.82 & 49.62 \\
0.4 & 98.56 & 56.71 & 98.56 & 86.26 & 96.40 & 51.50 & 88.93 & 48.85 \\
0.6 & 98.56 & 61.97 & 100.00 & 89.22 & 97.12 & 49.94 & 97.12 & 28.12 \\
0.8 & 98.56 & 64.48 & 100.00 & 92.36 & 99.28 & 45.77 & 98.56 & 39.72 \\
1.0 & 98.56 & 66.64 & 100.00 & 89.76 & 100.00 & 49.32 & 98.56 & 50.10 \\
\bottomrule
\end{tabular}
\vspace{-0.2in}
\end{table}

\section{Detection Methodology}
\label{sec:detection}
In this section, we present our methodology for detecting adversarial intermediate features in noisy collaborative inference systems. Given the integrity-targeted attack and the presence of communication noise described earlier, the objective is to design a detection framework that can distinguish adversarial manipulations from benign noise-induced deviations. To prevent adversarial features from propagating through inference stages and triggering unsafe decisions, detection must be performed at the point of feature reception. Accordingly, our method focuses on modeling the behavior of noise to more reliably separate benign perturbations from adversarial manipulations.

\subsection{Adversarial Variational Autoencoder (adVAE) Structure}

adVAEs extend the classical 
VAE framework by introducing adversarial 
objectives that make the latent space more discriminative for anomaly detection. VAEs~\cite{kingma2013auto} learn probabilistic latent representations through reconstruction, while GANs
use adversarial mechanisms to strengthen the model’s ability to detect distributional differences. adVAE combines these ideas by embedding a self-adversarial mechanism inside the VAE structure~\cite{wang2020advae}.

The model consists of three neural components: an encoder $E$, a generator (decoder) $G$, and a Gaussian transformer $T$ that perturbs latent 
representations. Given a normal input sample, the encoder produces a latent vector $z$, and the transformer applies a learned perturbation that effectively shifts the latent code's mean and variance to generate an adversarial latent vector $z_t$ that imitates anomalous patterns. The generator reconstructs samples from both $z$ and $z_t$, encouraging the model to amplify the distinction between normal and perturbed latent variables. In this self-adversarial structure, the transformer attempts to produce perturbed latent codes that remain difficult to distinguish from normal ones, while the generator emphasizes differences between the resulting reconstructions. Together, these interactions shape a latent space in which abnormal inputs naturally result in larger latent shifts and higher reconstruction errors.

\subsection{Motivation for Detection Features}
The misclassification attack used in this work is generated through a VAE model. Because of this, benign intermediate features and adversarially generated features may look very similar in the raw feature space, leading to strong overlap between their distributions. This makes it difficult for traditional anomaly detectors, which operate directly on raw features, to separate the two classes reliably.

However, the attacker’s manipulation becomes more apparent when the features are analyzed through a VAE. Since the attacker creates adversarial samples by interpolating in latent space, the resulting latent vectors show a noticeable shift in comparison to the genuine latent codes. In addition, the adversarial samples do not reconstruct as well as the benign samples, which leads to larger reconstruction error. Generally, these two quantities (latent shift and reconstruction error) capture differences that raw features do not reveal. 

This suggests that detection should make use of the same VAE-based structure that the attacker relies on, rather than depending only on raw features. Therefore, our defense focuses on extracting latent shift and reconstruction error. For clarity, given an intermediate feature $h$, the trained VAE based encoder produces latent code $z = E_{\phi}(h)$ and the decoder outputs a reconstructed $h_r = G_{\theta}(z)$, so we define Reconstruction Error (RE):
    \[
        \mathrm{RE}(h) = \|\, h - h_r \,\|_2^2,
    \]
which is small for benign samples as the VAE has learned to reconstruct their structure, and it is larger for adversarial samples.
And, the Latent Shift (LS):
    \[
        \mathrm{LS}(h) = \|\, z - z_t \,\|_2,
    \]
which measures the deviation of the encoded latent representation from the expected benign latent behavior learned by the VAE.


\subsection{Impulsive Noise Effects on VAE Detection }
\label{lab:Signature}
In our system, as discussed earlier, the intermediate features vector $h \in \mathbf{R^d}$ generated by the end device is corrupted by additive communication noise and then transmitted to the edge server. Under the noise, corruption does not affect the data uniformly as the communication disturbances occur in bursts \cite{shongwey2014impulse}. The probability that a coordinate experiences an impulsive
corruption is $p_b \ll 1$, and the expected number of corrupted features is $p_b d$, which is smaller than the feature dimension $d$. This sparsity in corruption of intermediate features shows that statistics such as the mean or variance, which are sensitive to outliers, are not reliable for characterizing the behavior of noisy benign samples. 
Given an intermediate feature vector ${h} = (h_1, h_2, \dots, h_d)$ the mean is:
\[
\bar{h} = \frac{1}{d}\sum_{i=1}^d h_i
\]
If even a single coordinate is corrupted by a large impulsive value,
\(
h_j \leftarrow h_j + n_j,
\) the mean can be shifted arbitrarily far from its true value. Both detection scores (RE and LS) are based on aggregation over feature dimensions. The objective function of the VAE-based anomaly detector is to minimize the  difference between an input $h \in \mathbb{R}^d$ and its reconstruction produced by the decoder: 

\begin{align*}
\mathcal{L}_{\text{VAE}}
    &= \mathcal{L}_{\text{rec}}\big(h,\; G_{\theta}(z)\big)
       \;+\; \lambda\,\mathcal{L}_{\text{KL}}\big(E_{\phi}(h)\big) \\[4pt]
    &= \mathcal{L}_{\text{rec}}(h,\; h_r)
       \;+\; \lambda\,\mathcal{L}_{\text{KL}}(\mu,\sigma),
\end{align*}

where the encoder outputs latent parameters $(\mu, \sigma)$, a latent vector 
$z$ is sampled from the Gaussian distribution $\mathcal{N}(\mu,\sigma^2)$, and $h_r = G_{\theta}(z)$ denotes the reconstructed feature.

\[
\mathcal{L}_{\text{rec}}(h, h_r) = \|\, h - h_r \,\|_2^{2}
\]

\begin{align*}
\mathcal{L}_{\text{KL}}(\mu,\sigma)
    &= \mathrm{KL}\!\left( \mathcal{N}(\mu,\sigma^2)
       \;\|\; \mathcal{N}(0,I) \right) \\[4pt]
    &= \int \mathcal{N}(z;\mu,\sigma^2)
       \log \frac{\mathcal{N}(z;\mu,\sigma^2)}
                 {\mathcal{N}(z;0,I)} \; dz \\[4pt]
    &= \frac{1}{2} \sum_{i=1}^d
       \left( 1 + \log \sigma_i^2 - \mu_i^2 - \sigma_i^2 \right)
\end{align*}

Although the VAE is trained to model benign feature distributions, this loss formulation is sensitive to impulsive perturbations during inference. In particular, a single corrupted coordinate $h_j \leftarrow h_j + n_j$ can change the VAE loss in two ways.
First, although the VAE is trained on benign data that may contain mild noise, the training distribution is dominated by clean or small perturbation samples. As a result, the encoder-decoder pair learns to model the typical structure of the data rather than the full range of possible noise outcomes. During inference, the decoder cannot reconstruct out-of-distribution features, and the reconstruction term $\| h-h_r \|_2^2$ tends to become large, causing a noisy benign sample to appear adversarial.

Second, the impulsive noise also disrupts the latent variables produced by the encoder. During training, the encoder learns to map benign features to posterior distributions $q_\phi(z\mid h) = \mathcal{N}(\mu,\sigma^2)$ that are encouraged through the Kullback-Leibler (KL) regularization term to lie close to the standard normal prior $p(z) = \mathcal{N}(0, I)$. When a test sample contains a structured noise burst, the encoder outputs $(\mu,\sigma)$  that deviates from this benign region, increasing the KL divergence and producing a latent shift that resembles the effect of adversarial manipulation.

\subsection{Noise-Aware Feature Extraction}
Although RE and LS provide useful signals for distinguishing benign and adversarial samples, their reliability degrades in noisy environments. As discussed earlier, communication noise perturbs intermediate features in ways that can resemble adversarial manipulation, causing the distributions of benign and adversarial samples to overlap. When detection relies only on these two metrics, many noise-corrupted benign samples may be incorrectly flagged as adversarial.

\begin{figure}[!t]
    \centering    \includegraphics[width=0.58\columnwidth]{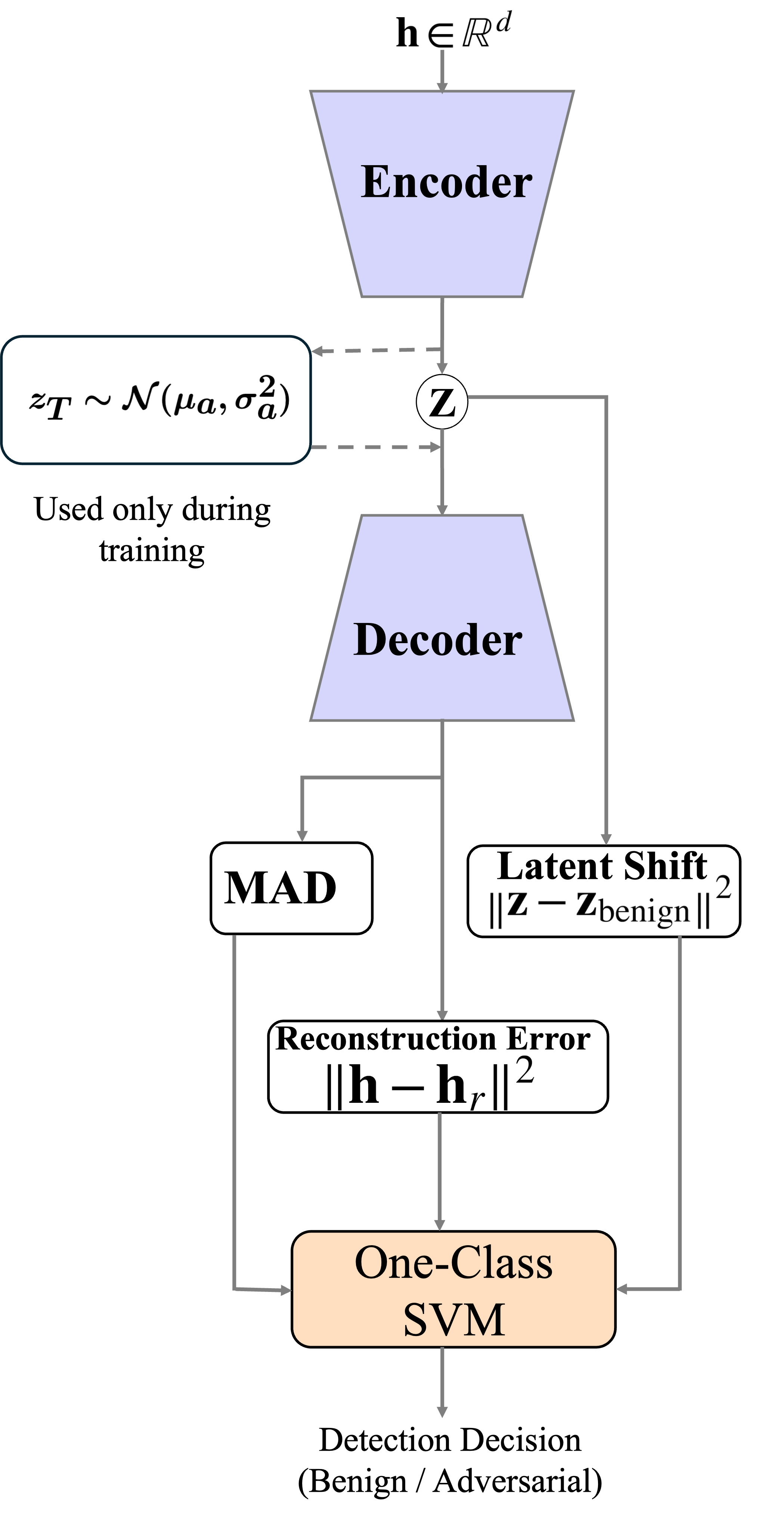}
    \caption{\scriptsize Core VAE-based structure underlying the proposed detection framework.}
    \label{fig:advae}
    \vspace{-0.2in}
\end{figure}

To address this limitation, we adopt a {\em semi–gray-box} modeling approach built on the VAE-based structure illustrated in Fig.~\ref {fig:advae}. Since the system has partial knowledge of the communication noise model, this information can be incorporated directly into the detection system. Rather than treating noise as an unspecified disturbance, we intentionally extract a statistical descriptor that captures its characteristic behavior. First, we analyze the residual vector from an input $h$ and its reconstructed form $h_{r} = G_{\theta}\!\big(E_{\phi}(h)\big)$ as follows:
\[
R(h) = |h - h_{r}|
\]
For intermediate features originating from a collaborative inference system, each coordinate of the residual indicates how much the reconstructed data deviates from the original input: 
\[
R(h) = \Bigl( 
\underbrace{\text{small}}_{\text{clean coords}},
\ldots,
\underbrace{\text{large spikes}}_{\text{noisy coords}}
\Bigr)
\]

As we explained in \ref{lab:Signature}, the environmental noise is impulsive and sparse; only a subset of the coordinates shows large deviation, and the rest of them remain close to the clean values. Therefore, in our system, a detection framework trained on benign but possibly noisy data must learn the typical structure of such residuals to be robust to noise spikes. To obtain a stable explanation of benign residual behavior, we adopt the Median Absolute Deviation (MAD), which is a well-known robust measure of scale. We do not use the median alone because it captures only the central value and not the magnitude of corruption. In contrast, MAD measures how much the values vary while remaining resistant to outliers:

\[
\mathrm{MAD}(h)
= \operatorname{median}\!\left( \left| R(h)_i - \operatorname{median}(R(h)) \right| \right)
\]
This robustness is important in our setting. Because MAD has a breakdown point of 50\%, more than half of the coordinates must be corrupted before the statistic is significantly affected.
This additional noise-aware feature complements latent shift and reconstruction error, enabling the detection framework to distinguish true adversarial manipulations from benign noise-induced perturbations.

\begin{figure*}[!t]
    \centering
    \includegraphics[width=0.90\textwidth]{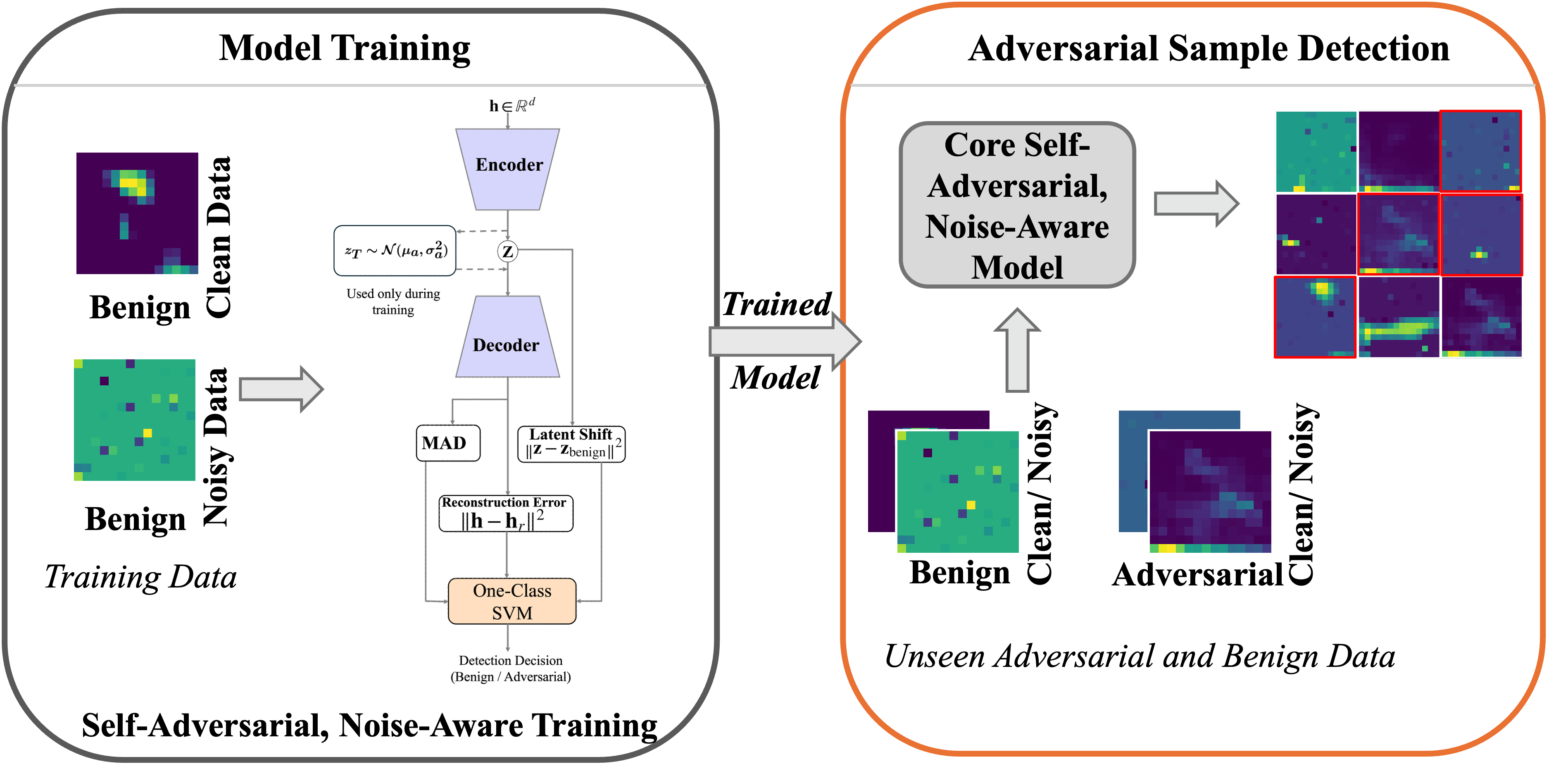}
    \caption{\scriptsize End-to-End Overview of the proposed training and detection pipeline.}
    \label{fig:pipeline}
    \vspace{-0.2in}
\end{figure*}

\subsection{Overall Detection Process}

The overall detection process for identifying adversarial misclassification attacks in a collaborative inference system is illustrated in Fig.~\ref{fig:pipeline}, building on the core VAE-based detection framework shown in Fig.~\ref {fig:advae}. We assume that, during an initial observation period, the system is not under attack. This allows the system to collect a set of benign (but potentially noisy) intermediate feature vectors transmitted from the end device to the edge server. Using this benign dataset, the system trains the adVAE model and extracts four detection features from each sample: reconstruction error, latent shift, residual median, and residual MAD. These features capture both VAE-based discrepancies and the characteristic structure of benign noise, as described in the previous subsections. Because the system has access only to benign data during training, a One-Class SVM (OC-SVM) is trained on the benign feature set. OC-SVM learns a compact boundary around the normal feature distribution without requiring adversarial examples, and labels any sample falling outside this region as anomalous.

Progressively, the adversary generates adversarial intermediate features following the threat model described in Section~\ref{sec:threat}. These manipulated features are transmitted through the noisy communication channel, resulting in a mixed stream of benign and adversarial (both noisy) sample. Each incoming feature vector is passed through the trained adVAE model to compute the detection feature set, which is then evaluated by the trained OC-SVM. 

Ideally, samples flagged as anomalous are blocked and not forwarded to the
edge server for processing by the remaining layers of the partitioned DNN. By preventing the server from processing adversarial samples, the system saves computational resources and reduces latency. As a result, the proposed detection pipeline prevents unsafe model decisions while improving overall system efficiency. 

\section{ Evaluation and Results}
\label{sec:evaluation}
Here, we evaluate the effectiveness of the proposed detection framework in distinguishing adversarial intermediate features from benign ones under realistic communication noise conditions. Our evaluation focuses on three key aspects: (i) the ability of the detection framework to separate benign noisy samples from adversarially generated samples, (ii) the influence of communication noise on overall detection performance, and (iii) the improvement in robustness achieved by incorporating the proposed noise-aware features alongside the VAE-based indicators.
The experimental study considers an intermediate-feature attack model that generates manipulated representations before transmission to the edge server. The proposed detection operates on deviations in representation space and does not rely on attack-specific signatures or internal details of the attack generation process.

\subsection{Scope of detection evaluation}
The results presented in Section~\ref{sec:threat} demonstrate that attacks launched from deeper cut-point layers achieve substantially higher success rates and induce misclassification with higher confidence. As the perturbation strength $\nu$ increases, attacks with $\nu > 0.6$ consistently yield stealthy misclassification across multiple architectures. Since these settings correspond to the worst-case adversarial threat scenarios, we restrict our evaluation to high attack strength to measure the effectiveness of the proposed noise-aware detection approach.

\subsection{Experiment setup}
We begin by outlining the used DNN models, the adVAE
architecture, the dataset, and the training details.

\noindent\underline{\textit{DNN Model:}}
We evaluate the proposed detection framework using VGG19, AlexNet, and MobileNet, which are the same downstream models considered in the prior misclassification attack study. These models are used to generate intermediate feature representations that are subsequently manipulated by the adVAE-based attack mechanism. This design allows us to directly assess the effectiveness of the proposed detection framework against adversarial samples produced under the misclassification attack scenario.

\noindent\underline{\textit{adVAE Architecture:}}
The proposed noise-aware detection framework is implemented in PyTorch and is based on an adVAE designed to model the distribution of benign intermediate features and to detect deviations caused by adversarial manipulation. For each model–layer configuration, a separate adVAE is trained due to differences in feature dimensionality. Given an intermediate tensor $x \in \mathbb{R}^{H \times W \times C}$ (e.g., $14 \times 14 \times 512$ for VGG19 layer 20), the input is flattened and passed through a two-layer fully connected encoder with ReLU activations. The latent dimension is fixed to 128. In addition to the standard encoder-decoder, a Gaussian latent transformer is included to apply a learnable mapping to the latent representation to enable sensitivity to latent deviations. The model is trained exclusively on benign intermediate features using a loss function that combines mean squared error (MSE) reconstruction loss and KL divergence regularization. The resulting reconstruction error and latent deviation statistics are passed to a downstream OC-SVM classifier for final anomaly detection. 


\noindent\underline{\textit{Dataset:}}
We use the CIFAR-100 dataset to train the DNN models for generating adversarial intermediate feature representations. CIFAR-100 contains 100 classes with 600 images per class, totaling 60,000 images. Following standard practice, 50,000 images (83.33\%) are used for training and 10,000 images (16.67\%) for testing. All images are RGB with a resolution of 32×32 pixels \cite{krizhevsky2009learning}. For the detection experiments, we evaluate performance on a subset of 4,000 samples drawn from the CIFAR-100 test set, consisting of 2,800 benign and 1,200 adversarial intermediate features. This reduced evaluation set is chosen to reflect realistic data availability in collaborative inference systems, where detection operates under limited and streaming data rather than large-scale offline evaluation.

\noindent\underline{\textit{Evaluation Metrics:}}
We evaluate the effectiveness of the detection framework in terms of metrics that jointly capture detection accuracy, robustness to class imbalance, and the unequal cost of detection errors, where missing attacks are significantly more costly than raising false alarms. Let $TP$, $TN$, $FP$, $FN$ denote the number of true positives, true negatives, false positives, and false negatives, respectively, where the positive class corresponds to adversarial intermediate features. 
\begin{itemize}
    \item \textit{Accuracy} gives a general idea of performance, but it can be misleading in anomaly detection setting where benign samples dominate. 
    \[
    \text{Accuracy} = \frac{TP + TN}{TP + TN + FP + FN}.
    \]
    \item \textit{Adversarial-class F1-score} provides the detection performance of adversarial samples. 
    \[
F1_{\text{anom}} =
\frac{2 \cdot \text{Precision}_{\text{anom}} \cdot \text{Recall}_{\text{anom}}}
{\text{Precision}_{\text{anom}} + \text{Recall}_{\text{anom}}}
\]

    \item \textit{Balanced accuracy} is used to handle class imbalance.
     \[
    \text{Balanced Accuracy} =
    \frac{1}{2}
    \left(
    \frac{TP}{TP+FN} + \frac{TN}{TN+FP}
    \right).
    \]
    
    \item \textit{Area Under the ROC Curve (AUROC)} measures the separability between benign and adversarial feature distributions.
\end{itemize}

\begin{figure}[t]
    \centering

    \includegraphics[width=\linewidth]{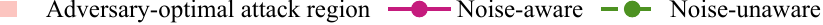}
    \vspace{-0.6em}

    \begin{subfigure}[t]{0.32\linewidth}
        \centering
        \includegraphics[width=\linewidth]{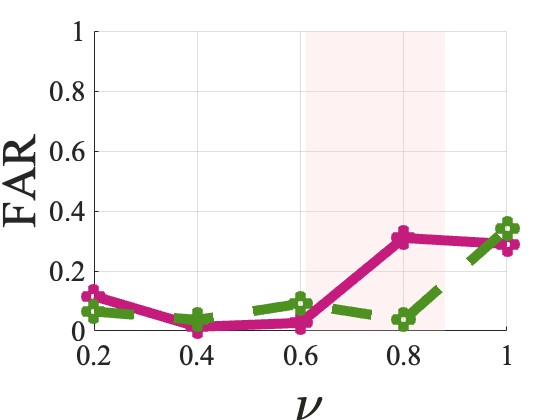}
        \label{fig:fpr_mod}
    \end{subfigure}\hfill
    \begin{subfigure}[t]{0.32\linewidth}
        \centering
        \includegraphics[width=\linewidth]{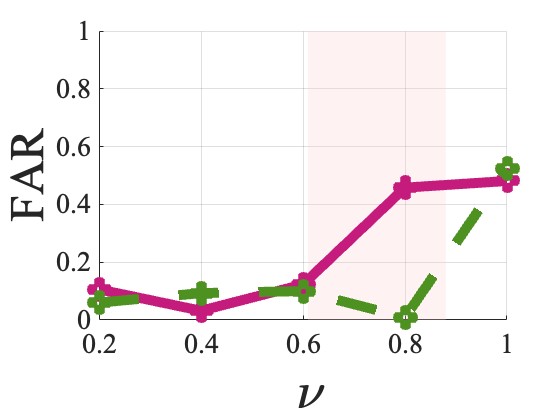}
        \label{fig:fpr_sev}
    \end{subfigure}\hfill
    \begin{subfigure}[t]{0.32\linewidth}
        \centering
        \includegraphics[width=\linewidth]{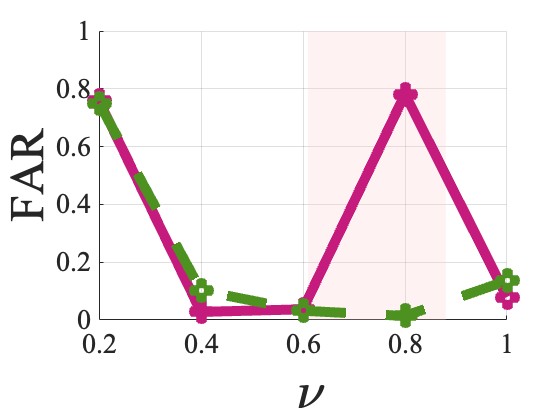}
        \label{fig:fpr_ext}
    \end{subfigure}

    \vspace{-0.9em}

    \begin{subfigure}[t]{0.32\linewidth}
        \centering
        \includegraphics[width=\linewidth]{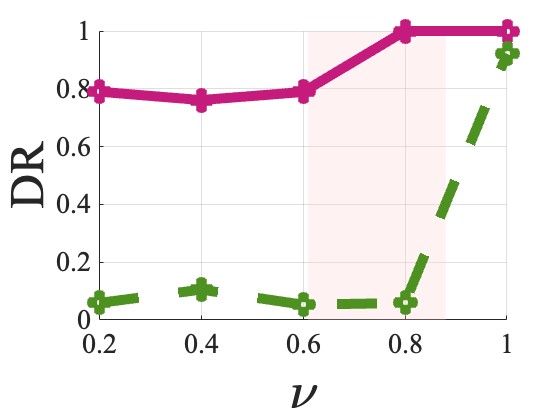}
        \caption{\small Moderate noise}
        \label{fig:tpr_mod}
    \end{subfigure}\hfill
    \begin{subfigure}[t]{0.32\linewidth}
        \centering
        \includegraphics[width=\linewidth]{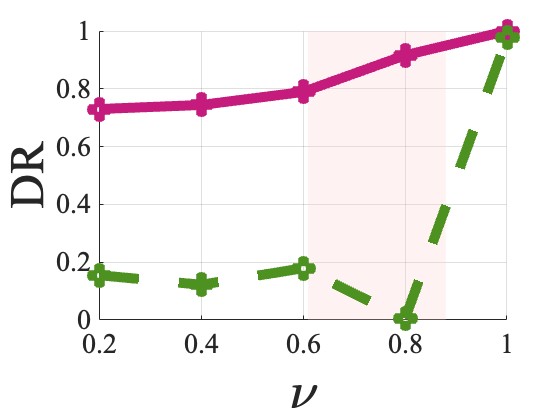}
        \caption{\small Severe noise}
        \label{fig:tpr_sev}
    \end{subfigure}\hfill
    \begin{subfigure}[t]{0.32\linewidth}
        \centering
        \includegraphics[width=\linewidth]{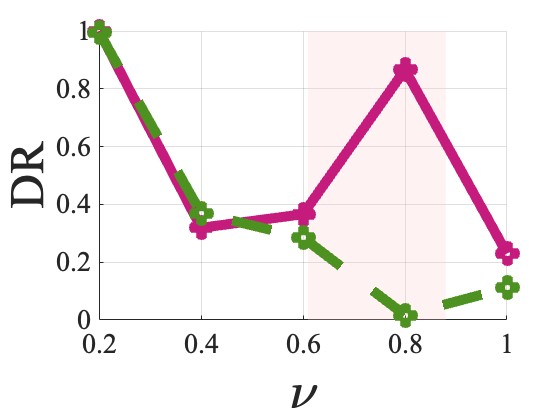}
        \caption{\small Extreme noise }
        \label{fig:tpr_ext}
    \end{subfigure}

    \caption{\scriptsize False alarm rate (FAR in top row) and detection rate (DR in bottom row) versus attack strength $\nu$ for VGG19 (layer~20) under increasing noise intensity.}
    \label{fig:fpr_tpr_alpha_vgg}
\end{figure}

\begin{table}[t]
\centering
\scriptsize
\setlength{\tabcolsep}{3.4pt}
\renewcommand{\arraystretch}{1.15}
\caption{\scriptsize VGG19 (Layer~20) detection performance under moderate noise. NA: Noise-aware; NU: Noise-unaware.}
\label{tab:vgg20_moderate}
\begin{tabular}{c|cc|cc|cc|cc}
\hline
\multirow{2}{*}{$\nu$} &
\multicolumn{2}{c|}{Accuracy} &
\multicolumn{2}{c|}{$F1_{\text{anom}}$} &
\multicolumn{2}{c|}{Balanced accuracy} &
\multicolumn{2}{c}{AUROC} \\
\cline{2-9}
 & NA & NU & NA & NU & NA & NU & NA & NU \\
\hline
0.2 & \textbf{0.835} & 0.476 & \textbf{0.834} & 0.105 & \textbf{0.837} & 0.479 & \textbf{0.824} & 0.348 \\
0.4 & \textbf{0.868} & 0.513 & \textbf{0.858} & 0.185 & \textbf{0.814} & 0.548 & \textbf{0.819} & 0.370 \\
0.6 & \textbf{0.875} & 0.456 & \textbf{0.870} & 0.093 & \textbf{0.881} & 0.481 & \textbf{0.834} & 0.685 \\
0.8 & \textbf{0.852} & 0.489 & \textbf{0.877} & 0.108 & \textbf{0.845} & 0.511 & \textbf{0.815} & 0.443 \\
1.0 & \textbf{0.867} & 0.800 & \textbf{0.890} & 0.833 & \textbf{0.855} & 0.790 & \textbf{0.857} & 0.730 \\
\hline
\end{tabular}
\vspace{-0.2in}
\end{table}



\subsection{Results}
For results, we present a detailed analysis of detection performance on VGG19 by varying the attack strength, noise intensity, and cut-point layer to examine how these factors affect the separability between benign and adversarial samples. Then we focus on the most effective attack setting, in terms of misclassification confidence, and provide summary results for AlexNet and MobileNet, under different noise intensities and cut-point configurations.   

Fig. \ref{fig:fpr_tpr_alpha_vgg} illustrates the trade-off between false alarm rate (FAR in top row) and detection rate (DR in bottom row) for VGG19 (layer~20) under increasing noise intensity. For small values of the attack strength~$\nu$, adversarial intermediate features remain close to benign representations, making them difficult to distinguish from noisy but legitimate samples. In this case, the noise-unaware baseline shows reduced sensitivity to weak attacks, resulting in a low FAR on benign samples at the cost of a severe drop in anomaly DR, as many weak adversarial samples are missed.

In contrast, by explicitly modeling noise variability, the noise-aware detection framework remains sensitive to structured deviations introduced by the attack and maintains high detection rates across attack strengths. As~$\nu$ increases—particularly in the highlighted adversary, optimal region where attacks induce confident misclassification, the adversarial features deviate more substantially from benign ones, enabling the proposed approach to achieve a more balanced trade-off between false alarms and missed detections. This balance is critical in collaborative inference, where high false alarm rates degrade overall system efficiency, while low detection rates allow adversarial intermediate features to be processed at the edge, wasting computational resources and posing potential security risks.

Tab. \ref{tab:vgg20_moderate} presents the detection performance of the proposed method for VGG19 (layer~20) under moderate noise. For small values of $\nu$, the adversarial intermediate features are only weakly perturbed and remain close to benign samples. In this case, the noise-unaware (NU) baseline achieves a lower false alarm rate primarily because it misses adversarial samples, resulting in near-random balanced accuracy and a very low anomaly F1-score. By explicitly modeling noise variability, the proposed noise-aware (NA) detection remains sensitive to deviations caused by adversarial manipulation. Even when the perturbation strength is small, the NA detection can identify anomalous samples under moderate noise conditions. As the attack strength increases and adversarial features deviate more substantially from benign ones, the consistently higher AUROC values are achieved by the proposed NA detection approach. This indicates more reliable class separability across attack strengths in noisy environments, where false alarms and missed detections directly impact system reliability.

\begin{table}[t]
\centering
\scriptsize
\setlength{\tabcolsep}{3.4pt}
\renewcommand{\arraystretch}{1.15}
\caption{\scriptsize VGG19 (Layer~20) detection performance under severe noise. NA: Noise-aware; NU: Noise-unaware.}
\label{tab:vgg20_severe}
\begin{tabular}{c|cc|cc|cc|cc}
\hline
\multirow{2}{*}{$\nu$} &
\multicolumn{2}{c|}{Accuracy} &
\multicolumn{2}{c|}{$F1_{\text{anom}}$} &
\multicolumn{2}{c|}{Balanced accuracy} &
\multicolumn{2}{c}{AUROC} \\
\cline{2-9}
 & NA & NU & NA & NU & NA & NU & NA & NU \\
\hline
0.2 &
\textbf{0.811} & 0.529 &
\textbf{0.803} & 0.257 &
\textbf{0.814} & 0.803 &
\textbf{0.820} & 0.548 \\

0.4 &
\textbf{0.850} & 0.496 &
\textbf{0.839} & 0.202 &
\textbf{0.856} & 0.515 &
\textbf{0.744} & 0.567 \\

0.6 &
\textbf{0.830} & 0.513 &
\textbf{0.831} & 0.282 &
\textbf{0.833} & 0.539 &
\textbf{0.853} & 0.603 \\

0.8 &
\textbf{0.738} & 0.474 &
\textbf{0.786} & 0.008 &
\textbf{0.729} & 0.497 &
\textbf{0.555} & 0.314 \\

1 &
\textbf{0.779} & 0.747 &
\textbf{0.830} & 0.807 &
\textbf{0.759} & 0.726 &
\textbf{0.657} & 0.647 \\
\hline
\end{tabular}
\end{table}

Tab. \ref{tab:vgg20_severe} presents the detection performance for VGG19 (layer~20) under severe noise conditions. Compared to the moderate noise setting, detection becomes more challenging as stronger noise increases the overlap between benign and adversarial intermediate features. We observe that the proposed noise-aware (NA) detection consistently outperforms the noise-unaware (NU) baseline across all attack strengths. At low attack strengths ($\nu$ = 0.2, 0.4), adversarial perturbation is small and remains masked by high noise intensity. As a result, the NU detection misses most adversarial samples and achieves near-random performance as reflected by F1-score and AUROC values. In contrast, the proposed NA detection remains more sensitive by explicitly modeling noise characteristics. As the attack strength increases, adversarial manipulations become more pronounced and easier to distinguish from benign noise, resulting in improved performance for both methods.

\begin{table}[t]
\centering
\scriptsize
\setlength{\tabcolsep}{3.4pt}
\renewcommand{\arraystretch}{1.15}
\caption{\scriptsize VGG19 (Layer~20) detection performance under extreme noise. NA: Noise-aware; NU: Noise-unaware.}
\label{tab:vgg20_extreme}
\begin{tabular}{c|cc|cc|cc|cc}
\hline
\multirow{2}{*}{$\nu$} &
\multicolumn{2}{c|}{Accuracy} &
\multicolumn{2}{c|}{$F1_{\text{anom}}$} &
\multicolumn{2}{c|}{Balanced accuracy} &
\multicolumn{2}{c}{AUROC} \\
\cline{2-9}
 & NA & NU & NA & NU & NA & NU & NA & NU \\
\hline
0.2 &0.637 &0.637 &0.742 &0.740&0.618 &0.618 &0.601 & 0.566\\
0.4 &0.630&0.621 & 0.506& 0.475&0.646 &0.639 &0.685 &0.639 \\
0.6 &0.647 &0.606 &0.523 & 0.433& 0.665&0.626 & 0.689&0.641 \\
0.8 & 0.557& 0.478&0.672 &0.033 &0.542 &0.501 & 0.465& 0.434\\
1 & 0.547&0.457 &0.355 & 0.181&0.575&0.488& 0.594&0.510 \\
\hline
\end{tabular}
\vspace{-0.2in}
\end{table}

Tab. \ref{tab:vgg20_extreme} presents the detection performance of VGG19 (layer 20) under extreme noise conditions. Under this setting, communication noise dominates the intermediate features, leading to substantial overlap between benign and adversarial feature distributions. This makes the separation between benign and adversarial samples difficult. Unlike the moderate and severe noise cases, the NU baseline shows improvement in accuracy. However, this is due to classifying most samples as benign which does not necessarily indicate better anomaly detection, as the F1-score remains low. As a result, detection performance degrades for both methods compared to the moderate and severe noise scenarios.

Based on the threat analysis, attacks with $\nu > 0.6$ represent the most challenging and safety-critical case. Accordingly, we focus on a representative high-strength setting with $\nu = 0.8$ to concentrate the evaluation on the most informative attack conditions. For VGG19, Tab. \ref{tab:vgg_layers_severe} presents results for additional cut-point layers (layers 12, 16, and 18) evaluated under severe noise only, since moderate noise is not sufficiently challenging, and extreme noise causes substantial overlap between benign and adversarial features. Based on the attack results, we observed that the proposed NA detection maintains strong detection performance at earlier cut-point layers. This behavior is consistent with the corresponding attack results, which show that adversarial samples generated from earlier layers exhibit lower misclassification confidence. Lower confidence means that the intermediate adversarial features deviate substantially from the benign feature distribution. As the cut-point moves deeper, the attack becomes increasingly stealthy and reduces the separation between benign and adversarial samples in the feature space used for detection. The NU baseline shows the same trend across layers; however, due to the lack of explicit noise modeling, its performance degradation is more pronounced.


\begin{table}[t]
\centering
\scriptsize
\setlength{\tabcolsep}{3.8pt}
\renewcommand{\arraystretch}{1.15}
\caption{\scriptsize Detection performance across VGG19 cut-point layers under severe noise with attack strength $\nu=0.8$. NA: Noise-aware; NU: Noise-unaware.}
\label{tab:vgg_layers_severe}
\begin{tabular}{c|cc|cc|cc|cc}
\hline
\multirow{2}{*}{Layer} &
\multicolumn{2}{c|}{Accuracy} &
\multicolumn{2}{c|}{$F1_{\text{anom}}$} &
\multicolumn{2}{c|}{Balanced accuracy} &
\multicolumn{2}{c}{AUROC} \\
\cline{2-9}
 & NA & NU & NA & NU & NA & NU & NA & NU \\
\hline
12 & \textbf{0.991} & 0.749 & \textbf{0.991} & 0.688 & \textbf{0.992} & 0.742 & \textbf{1.000} & 0.794 \\
16 &\textbf{0.984} & 0.669 & \textbf{0.984} &0.561 & \textbf{0.984} & 0.669 & 1 & 0.750 \\
18 & \textbf{0.990} & 0.498 & \textbf{0.990} & 0.000 & \textbf{0.990} & 0.480 & \textbf{0.994} & 0.662 \\
\hline
\end{tabular}
\end{table}

Tab. \ref{tab:alexnet_noise} reports the detection performance for AlexNet across different cut-point layers under a fixed near-optimal attack strength ($\nu = 0.8$) while varying the noise intensity. The results demonstrate a strong layer-dependent behavior that is highly correlated with the effectiveness of the attack at each layer. As shown in Tab.~\ref{tab:alexnet_attack}, adversarial samples generated from earlier layers exhibit lower misclassification confidence, indicating that these samples deviate more substantially from the benign feature distribution. This larger deviation makes detection easier, resulting in higher detection performance (e.g., accuracy, anomaly F1-score, and AUROC) for earlier layers. As the cut-point moves to deeper layers, the attack confidence increases, indicating that adversarial features become more aligned with the benign data manifold, which makes detection more challenging. Under moderate noise, the proposed detection outperforms the baseline; however, as the noise intensifies, both detections' performance degrades. 

\begin{table}[t]
\centering
\scriptsize
\setlength{\tabcolsep}{3.2pt}
\renewcommand{\arraystretch}{1.12}
\caption{\scriptsize AlexNet detection performance across cut-point layers under varying noise intensity with attack strength $\nu=0.8$. NA: Noise-aware; NU: Noise-unaware.}
\label{tab:alexnet_noise}
\begin{tabular}{cc|cc|cc|cc|cc}
\hline
\multirow{2}{*}{Noise} & \multirow{2}{*}{Layer} &
\multicolumn{2}{c|}{Accuracy} &
\multicolumn{2}{c|}{$F1_{\text{anom}}$} &
\multicolumn{2}{c|}{Balanced acc.} &
\multicolumn{2}{c}{AUROC} \\
\cline{3-10}
 &  & NA & NU & NA & NU & NA & NU & NA & NU \\
\hline

\multirow{4}{*}{Moderate}
& 3  & 0.891 & 0.641 & 0.886 & 0.624 & 0.891 & 0.642 & 0.916 & 0.705 \\
& 6  & 0.836 & 0.677 & 0.808 & 0.664 & 0.832 & 0.677 & 0.733 & 0.636 \\
& 8  & 0.847 & 0.769 & 0.836 & 0.782 & 0.853 & 0.768 & 0.808 & 0.815 \\
& 10 & \textbf{0.820} & 0.609 & \textbf{0.821} & 0.578 & \textbf{0.820} & 0.612 & \textbf{0.770} & 0.592 \\
\hline

\multirow{4}{*}{Severe}
& 3  & 0.897 & 0.510 & 0.902 & 0.417 & 0.897 & 0.512 & 0.953 & 0.534 \\
& 6  & 0.730 & 0.572 & 0.667 & 0.499 & 0.725 & 0.568 & 0.768 & 0.606 \\
& 8  & 0.845 & 0.565 & 0.861 & 0.531 & 0.840 & 0.571 & 0.918 & 0.616 \\
& 10 & \textbf{0.585} & 0.564 & \textbf{0.533} & 0.521 & \textbf{0.588} & 0.567 & 0.567 & \textbf{0.577} \\
\hline

\multirow{4}{*}{Extreme}
& 3  & 0.643 & 0.460 & 0.511 & 0.331 & 0.648 & 0.463 & 0.705 & 0.489 \\
& 6  & 0.590 & 0.505 & 0.331 & 0.371 & 0.579 & 0.499 & 0.528 & 0.498 \\
& 8  & 0.544 & 0.559 & 0.311 & 0.493 & 0.566 & 0.568 & 0.625 & 0.587 \\
& 10 & \textbf{0.536} & 0.489 & \textbf{0.514} & 0.375 & \textbf{0.537} & 0.494 & \textbf{0.527} & 0.492 \\
\hline

\end{tabular}
\vspace{-0.2in}
\end{table}

\begin{table}[t]
\centering
\scriptsize
\setlength{\tabcolsep}{3.6pt}
\renewcommand{\arraystretch}{1.15}
\caption{\scriptsize Detection performance for MobileNet at different cut-point layers under varying noise intensities with attack strength $\nu=0.8$. NA: Noise-aware; NU: Noise-unaware.}
\label{tab:mobilenet_noise}
\begin{tabular}{c|c|cc|cc|cc|cc}
\hline
\multirow{2}{*}{Noise} & 
\multirow{2}{*}{Layer} &
\multicolumn{2}{c|}{Accuracy} &
\multicolumn{2}{c|}{$F1_{\text{anom}}$} &
\multicolumn{2}{c|}{Balanced accuracy} &
\multicolumn{2}{c}{AUROC} \\
\cline{3-10}
 &  & NA & NU & NA & NU & NA & NU & NA & NU \\
\hline
\multirow{3}{*}{Moderate}
 & \textbf{20} & \textbf{0.831} & \textbf{0.520} & \textbf{0.811} & \textbf{0.211} & \textbf{0.831} & \textbf{0.521} & \textbf{0.838} & \textbf{0.481} \\
 & 40 & 0.829 & 0.793 & 0.844 & 0.817 & 0.834 & 0.799 & 0.830 & 0.817 \\
 & 50 & 0.849 & 0.670 & 0.836 & 0.538 & 0.848 & 0.665 & 0.945 & 0.726 \\
\hline
\multirow{3}{*}{Severe}
 & \textbf{20} & \textbf{0.745} & \textbf{0.533} & \textbf{0.663} & \textbf{0.243} & \textbf{0.746} & \textbf{0.534} & \textbf{0.819} & \textbf{0.542} \\
 & 40 & 0.927 & 0.855 & 0.928 & 0.843 & 0.929 & 0.853 & 0.981 & 0.918 \\
 & 50 & 0.854 & 0.866 & 0.847 & 0.857 & 0.853 & 0.865 & 0.926 & 0.928 \\
\hline
\multirow{3}{*}{Extreme}
 & \textbf{20} & \textbf{0.587} & \textbf{0.503} & \textbf{0.341} & \textbf{0.147} & \textbf{0.588} & \textbf{0.504} & \textbf{0.619} & \textbf{0.473} \\
 & 40 & 0.897 & 0.791 & 0.888 & 0.756 & 0.896 & 0.786 & 0.943 & 0.863 \\
 & 50 & 0.841 & 0.788 & 0.829 & 0.761 & 0.840 & 0.786 & 0.914 & 0.871 \\
\hline
\end{tabular}
\end{table}

Tab. \ref{tab:mobilenet_noise} reports the detection performance for MobileNet across different cut-point layers under varying noise intensities with a $\nu = 0.8$. For MobileNet, detection performance is closely related to the success of the attack in terms of misclassification confidence, which reflects how stealthy the adversarial samples are. As presented in Tab. \ref{tab:MobileNet_attack}, layers where the attack achieves higher confidence produce more stealthy adversarial features that are closer to the benign data distribution. In these layers, detection performance is lower compared to layers where the attack confidence is smaller. However, noise affects this behavior by causing overlap between benign and adversarial samples. The impact of noise intensity on the proposed noise-aware detection is strongly layer-dependent. While detection performance remains stable for a deeper cut-point layer, a noticeable degradation is observed at layer 20 as noise intensity increases. This behavior can be attributed to the lower-level nature of features at earlier layers, which are more sensitive to noise 

\begin{table}[t]
\centering
\footnotesize
\setlength{\tabcolsep}{3pt}
\renewcommand{\arraystretch}{0.95}
\caption{Per-sample detection latency at the edge for different cut layers and architectures.}
\begin{tabular}{l|c|c|c}
\toprule
\textbf{Model} & \textbf{Cut Layer} & \textbf{Output Size} & \textbf{Latency (ms)} \\
\midrule
\multirow{4}{*}{VGG19}
& 12 & 28×28×512 & 46.84 \\
& 16 & 14×14×512 & 13.25 \\
& 18 & 14×14×512 & 13.22 \\
& 20 & 14×14×512 & 13.09 \\
\midrule
\multirow{4}{*}{AlexNet} 
& 3  & 27×27×192 & 17.41 \\
& 6  & 13×13×384 & 8.82 \\
& 8  & 13×13×256 & 6.53 \\
& 10 & 13×13×256 & 6.50 \\
\midrule
\multirow{4}{*}{MobileNet}
& 20 & 56×56×128 & 46.90 \\
& 40 & 14×14×512 & 12.87 \\
& 50 & 14×14×512 & 12.86 \\
& 63 & 14×14×512 & 12.88 \\
\bottomrule
\end{tabular}
\label{tab:latency}
\vspace{-0.2in}
\end{table}


Finally, we evaluate the detection inference latency on a Dell PowerEdge server running Ubuntu Linux. The system is equipped with an Intel Xeon Gold 6240R CPU @ 2.40 GHz (48 logical cores), 64 GB RAM, and an NVIDIA RTX A4000 GPU. Results summarized in Tab.  \ref{tab:latency} show that the per-sample detection latency remains low across different architectures and cut layers. Since training is performed offline, the reported latency reflects only the online inference stage at deployment. This includes the adVAE forward pass followed by OC-SVM classification.
These results demonstrate that the proposed detection mechanism introduces minimal computational overhead at deployment time, making it suitable for real-time DNN inference scenarios. All evaluation related codes and data are available through Github~\cite{git}.


\section{Discussions and Conclusions}
\label{sec:conclusion}
In this paper, we investigated the problem of detecting adversarial intermediate features in collaborative inference environments under realistic communication noise. We first analyzed how impulsive noise affects the distribution of benign intermediate features and how this effect can cause substantial overlap with adversarial feature distributions, making detection particularly challenging. To address this challenge, we proposed an online, semi–gray-box, noise-aware detection framework that operates on intermediate representations and explicitly incorporates prior knowledge about noise behavior. 
The proposed method combined VAE-based deviation measures with a robust noise-aware descriptor learned from statistical characteristics of benign noisy data. As a result, it could distinguish noise-corrupted benign samples from adversarial manipulations even when their distributions are highly overlapped.

We conducted extensive evaluations across multiple CNN architectures, different cut-point layers, varying attack strengths, and different noise intensities. The results demonstrated that the proposed detection framework consistently outperforms the noise-unaware baseline and successfully separates adversarial samples from benign ones in most practical scenarios. Our findings revealed that the detection performance is highly influenced by the cut-point position. Adversarial samples generated from earlier cut-point layers typically induce lower misclassification confidence in the downstream model, as they have larger structural deviations from benign features, making them easier to detect. As the cut-point layer moved deeper, adversarial features became more stealthy and resembled benign noisy representations. As a result, they exhibited higher distributional overlap and reduced separability. While both noise-aware and noise-unaware approaches followed this trend, the proposed noise-aware approach achieved higher detection performance.

\bibliographystyle{ieeetr}
\bibliography{reference}

@inproceedings{yousefi2024intent,
  title={Intent-Driven Data Falsification Attack on Collaborative IoT-Edge Environments},
  author={Yousefi, Shima and Bhattacharjee, Shameek and Debroy, Saptarshi},
  booktitle={2024 IEEE/ACM Symposium on Edge Computing (SEC)},
  pages={425--430},
  year={2024},
  organization={IEEE}
}

@article{zhang2025c2mec,
  title={c2mec: Cooperative Multi-split and Multi-hop Edge Computing Based on Deep Reinforcement Learning},
  author={Zhang, Xiaojie and Wang, Peng and Yousefi, Shima and Debroy, Saptarshi and Li, Keqin},
  journal={ACM Transactions on Embedded Computing Systems},
  year={2025},
  publisher={ACM New York, NY}
}

@article{mireshghallah2020principled,
  title={A principled approach to learning stochastic representations for privacy in deep neural inference},
  author={Mireshghallah, Fatemehsadat and Taram, Mohammadkazem and Jalali, Ali and Elthakeb, Ahmed Taha and Tullsen, Dean and Esmaeilzadeh, Hadi},
  journal={arXiv preprint arXiv:2003.12154},
  year={2020}
}

@inproceedings{yousefi2025advar,
  title={Advar-dnn: Adversarial misclassification attack on collaborative dnn inference},
  author={Yousefi, Shima and Mounesan, Motahare and Debroy, Saptarshi},
  booktitle={2025 IEEE 50th Conference on Local Computer Networks (LCN)},
  pages={1--9},
  year={2025},
  organization={IEEE}
}

@inproceedings{he2019model,
  title={Model inversion attacks against collaborative inference},
  author={He, Zecheng and Zhang, Tianwei and Lee, Ruby B},
  booktitle={Proceedings of the 35th annual computer security applications conference},
  pages={148--162},
  year={2019}
}

@article{he2020attacking,
  title={Attacking and protecting data privacy in edge--cloud collaborative inference systems},
  author={He, Zecheng and Zhang, Tianwei and Lee, Ruby B},
  journal={IEEE Internet of Things Journal},
  volume={8},
  number={12},
  pages={9706--9716},
  year={2020},
  publisher={IEEE}
}

@article{su2019one,
  title={One pixel attack for fooling deep neural networks},
  author={Su, Jiawei and Vargas, Danilo Vasconcellos and Sakurai, Kouichi},
  journal={IEEE Transactions on Evolutionary Computation},
  volume={23},
  number={5},
  pages={828--841},
  year={2019},
  publisher={IEEE}
}

@inproceedings{biggio2013evasion,
  title={Evasion attacks against machine learning at test time},
  author={Biggio, Battista and Corona, Igino and Maiorca, Davide and Nelson, Blaine and {\v{S}}rndi{\'c}, Nedim and Laskov, Pavel and Giacinto, Giorgio and Roli, Fabio},
  booktitle={Joint European conference on machine learning and knowledge discovery in databases},
  pages={387--402},
  year={2013},
  organization={Springer}
}

@inproceedings{man2020ghostimage,
  title={$\{$GhostImage$\}$: Remote perception attacks against camera-based image classification systems},
  author={Man, Yanmao and Li, Ming and Gerdes, Ryan},
  booktitle={23rd International Symposium on Research in Attacks, Intrusions and Defenses (RAID 2020)},
  pages={317--332},
  year={2020}
}

@article{goodfellow2014explaining,
  title={Explaining and harnessing adversarial examples},
  author={Goodfellow, Ian J and Shlens, Jonathon and Szegedy, Christian},
  journal={arXiv preprint arXiv:1412.6572},
  year={2014}
}

@article{guetta2021dodging,
  title={Dodging attack using carefully crafted natural makeup},
  author={Guetta, Nitzan and Shabtai, Asaf and Singh, Inderjeet and Momiyama, Satoru and Elovici, Yuval},
  journal={arXiv preprint arXiv:2109.06467},
  year={2021}
}

@article{liu2020noise,
  title={Noise removal in the presence of significant anomalies for industrial IoT sensor data in manufacturing},
  author={Liu, Yuehua and Dillon, Tharam and Yu, Wenjin and Rahayu, Wenny and Mostafa, Fahed},
  journal={IEEE Internet of Things Journal},
  volume={7},
  number={8},
  pages={7084--7096},
  year={2020},
  publisher={IEEE}
}

@article{bigdeli2017fast,
  title={A fast and noise resilient cluster-based anomaly detection},
  author={Bigdeli, Elnaz and Mohammadi, Mahdi and Raahemi, Bijan and Matwin, Stan},
  journal={Pattern Analysis and Applications},
  volume={20},
  number={1},
  pages={183--199},
  year={2017},
  publisher={Springer}
}

@article{li2019measurement,
  title={Measurement and characterization of electromagnetic noise in edge computing networks for the industrial Internet of Things},
  author={Li, Huiting and Liu, Liu and Li, Yiqian and Yuan, Ze and Zhang, Kun},
  journal={Sensors},
  volume={19},
  number={14},
  pages={3104},
  year={2019},
  publisher={MDPI}
}

@article{foi2008practical,
  title={Practical Poissonian-Gaussian noise modeling and fitting for single-image raw-data},
  author={Foi, Alessandro and Trimeche, Mejdi and Katkovnik, Vladimir and Egiazarian, Karen},
  journal={IEEE transactions on image processing},
  volume={17},
  number={10},
  pages={1737--1754},
  year={2008},
  publisher={IEEE}
}

@inproceedings{ma2012performance,
  title={Performance analysis of 3-D data transmission system in the presence of impulsive noise},
  author={Ma, Shuang and Chen, Zhenxing and Kang, Seog Geun},
  booktitle={2012 International Conference on ICT Convergence (ICTC)},
  pages={601--602},
  year={2012},
  organization={IEEE}
}

@article{kingma2013auto,
  title={Auto-encoding variational bayes},
  author={Kingma, Diederik P and Welling, Max},
  journal={arXiv preprint arXiv:1312.6114},
  year={2013}
}

@article{wang2020advae,
  title={adVAE: A self-adversarial variational autoencoder with Gaussian anomaly prior knowledge for anomaly detection},
  author={Wang, Xuhong and Du, Ying and Lin, Shijie and Cui, Ping and Shen, Yuntian and Yang, Yupu},
  journal={Knowledge-Based Systems},
  volume={190},
  pages={105187},
  year={2020},
  publisher={Elsevier}
}

@article{momeny2021noise,
  title={A noise robust convolutional neural network for image classification},
  author={Momeny, Mohammad and Latif, Ali Mohammad and Sarram, Mehdi Agha and Sheikhpour, Razieh and Zhang, Yu Dong},
  journal={Results in Engineering},
  volume={10},
  pages={100225},
  year={2021},
  publisher={Elsevier}
}

@inproceedings{shongwey2014impulse,
  title={On impulse noise and its models},
  author={Shongwey, Thokozani and Vinck, AJ Han and Ferreira, Hendrik C},
  booktitle={18th IEEE International Symposium on Power Line Communications and Its Applications},
  pages={12--17},
  year={2014},
  organization={IEEE}
}

@misc{krizhevsky2009learning,
  title={Learning multiple layers of features from tiny images.(2009)},
  author={Krizhevsky, Alex and Hinton, Geoffrey and others},
  year={2009}
}

@inproceedings{gan2025exploring,
  title={Exploring Transferability of Adversarial Examples from Resource-Constrained Devices},
  author={Gan, Houchao and Yousefi, Shima and Debroy, Saptarshi},
  booktitle={Proceedings of the Tenth ACM/IEEE Symposium on Edge Computing},
  pages={1--6},
  year={2025}
}

@article{wu2024deep,
  title={Deep learning solutions for smart city challenges in urban development},
  author={Wu, Pengjun and Zhang, Zhanzhi and Peng, Xueyi and Wang, Ran},
  journal={Scientific Reports},
  volume={14},
  number={1},
  pages={5176},
  year={2024},
  publisher={Nature Publishing Group UK London}
}

@article{lapidoth2020encoder,
  title={Encoder-assisted communications over additive noise channels},
  author={Lapidoth, Amos and Marti, Gian},
  journal={IEEE Transactions on Information Theory},
  volume={66},
  number={11},
  pages={6607--6616},
  year={2020},
  publisher={IEEE}
}

@article{qing2024detection,
  title={Detection of adversarial attacks via disentangling natural images and perturbations},
  author={Qing, Yuanyuan and Bai, Tao and Liu, Zhuotao and Moulin, Pierre and Wen, Bihan},
  journal={IEEE Transactions on Information Forensics and Security},
  volume={19},
  pages={2814--2825},
  year={2024},
  publisher={IEEE}
}

@article{feinman2017detecting,
  title={Detecting adversarial samples from artifacts},
  author={Feinman, Reuben and Curtin, Ryan R and Shintre, Saurabh and Gardner, Andrew B},
  journal={arXiv preprint arXiv:1703.00410},
  year={2017}
}

@article{mu2025robust,
  title={Robust Adversarial Example Detection Algorithm Based on High-Level Feature Differences},
  author={Mu, Hua and Li, Chenggang and Peng, Anjie and Wang, Yangyang and Liang, Zhenyu},
  journal={Sensors},
  volume={25},
  number={6},
  pages={1770},
  year={2025},
  publisher={MDPI}
}

@inproceedings{saha2025detection,
  title={Detection of Misreporting Attacks on Software-Defined Immersive Environments},
  author={Saha, Sourya and Absur, Md Nurul and Yousefi, Shima and Debroy, Saptarshi},
  booktitle={2025 21st International Conference on Network and Service Management (CNSM)},
  pages={1--7},
  year={2025},
  organization={IEEE}
}

@INPROCEEDINGS{inferedge,
  author={Mounesan, Motahare and Zhang, Xiaojie and Debroy, Saptarshi},
  booktitle={NOMS 2025-2025 IEEE Network Operations and Management Symposium}, 
  title={Infer-EDGE: Dynamic DNN Inference Optimization in Just-in-Time Edge-AI Implementations}, 
  year={2025},
  volume={},
  number={},
  pages={1-9},
  doi={10.1109/NOMS57970.2025.11073623}}

@INPROCEEDINGS{vec,
  author={Mounesan, Motahare and Lemus, Mauro and Yeddulapalli, Hemanth and Calyam, Prasad and Debroy, Saptarshi},
  booktitle={2024 IEEE 8th International Conference on Fog and Edge Computing (ICFEC)}, 
  title={Reinforcement Learning-driven Data-intensive Workflow Scheduling for Volunteer Edge-Cloud}, 
  year={2024},
  volume={},
  number={},
  pages={79-88},
  doi={10.1109/ICFEC61590.2024.00016}}

@inproceedings{xiaojie-sec,
author = {Zhang, Xiaojie and Debroy, Saptarshi},
title = {Adaptive task offloading over wireless in mobile edge computing},
year = {2019},
isbn = {9781450367332},
publisher = {Association for Computing Machinery},
address = {New York, NY, USA},
url = {https://doi.org/10.1145/3318216.3363328},
doi = {10.1145/3318216.3363328},
booktitle = {Proceedings of the 4th ACM/IEEE Symposium on Edge Computing},
pages = {323–325},
numpages = {3},
location = {Arlington, Virginia},
series = {SEC '19}
}

@INPROCEEDINGS{effect-dnn,
  author={Zhang, Xiaojie and Mounesan, Motahare and Debroy, Saptarshi},
  booktitle={2023 IEEE 24th International Symposium on a World of Wireless, Mobile and Multimedia Networks (WoWMoM)}, 
  title={EFFECT-DNN: Energy-efficient Edge Framework for Real-time DNN Inference}, 
  year={2023},
  volume={},
  number={},
  pages={10-20},
  doi={10.1109/WoWMoM57956.2023.00015}}

@inproceedings{mahboob2021coronavirus,
  title={A coronavirus herd immunity optimizer for intrusion detection system},
  author={Mahboob, Amir Soltany and Shahhoseini, Hadi Shahriar and Moghaddam, Mohammad Reza Ostadi and Yousefi, Shima},
  booktitle={2021 29th Iranian conference on electrical engineering (ICEE)},
  pages={579--585},
  year={2021},
  organization={IEEE}
}

@inproceedings{loodaricheh2026mage,
  title={MAGE-ID: A Multimodal Generative Framework for Intrusion Detection Systems},
  author={Loodaricheh, Mahdi Arab and Manshaei, Mohammad Hossein and Raja, Anita},
  booktitle={2026 IEEE International Conference on Computing, Networking and Communications (ICNC)},
  year={2026},
  pages={1--6},
  publisher={IEEE},
  doi={10.1109/ICNC68183.2026.11416962}
}

@misc{git,
  author      = {GitHub},
  year        = {2026},
  title       = {Github repository},
  note        = {Accessed: Mar 14, 2026},
  howpublished= {\url{https://github.com/dissectlab/Misclassification-CCGrid2026.git}}
}

\end{document}